\documentclass{article}

\PassOptionsToPackage{numbers}{natbib}
 \usepackage[preprint]{neurips_2026}

\usepackage[utf8]{inputenc} 
\usepackage[T1]{fontenc}    
\usepackage{hyperref}       
\usepackage{url}            
\usepackage{booktabs}       
\usepackage{amsfonts}       
\usepackage{nicefrac}       
\usepackage{microtype}      
\usepackage{xcolor}         

\usepackage{microtype}
\usepackage{graphicx}
\usepackage{subcaption}
\usepackage{booktabs} 

\usepackage{framed}
\usepackage{xcolor}
\definecolor{shadecolor}{gray}{0.95}

\usepackage{hyperref}

\usepackage{amsmath,amsfonts,bm}

\def\eqref#1{equation~\ref{#1}}

\def\1{\bm{1}}

\DeclareMathAlphabet{\mathsfit}{\encodingdefault}{\sfdefault}{m}{sl}
\SetMathAlphabet{\mathsfit}{bold}{\encodingdefault}{\sfdefault}{bx}{n}

\usepackage{amsmath, amsfonts}
\usepackage{algorithmic}
\usepackage{graphicx}
\usepackage{textcomp}
\usepackage{xcolor}
\usepackage{multirow}
\usepackage{mathtools}
\usepackage{tcolorbox}
\usepackage{amssymb}
\usepackage{capt-of}
\usepackage{enumitem}
\usepackage{makecell}
\usepackage{float}
\usepackage{fancyhdr}
\usepackage{colortbl}
\usepackage{arydshln}
\usepackage{xcolor,soul}
\usepackage{tcolorbox}
\usepackage{wrapfig}
\usepackage{titletoc} 
\usepackage[ruled,linesnumbered]{algorithm2e}
\RestyleAlgo{ruled} 
\usepackage{pifont} 
\usepackage{tikz}
\usepackage{booktabs}
\usepackage{listings}
\usepackage{natbib}
\definecolor{codegreen}{rgb}{0,0.6,0}
\definecolor{codegray}{rgb}{0.5,0.5,0.5}
\definecolor{codepurple}{rgb}{0.58,0,0.82}
\definecolor{backcolour}{rgb}{0.95,0.95,0.92}

\lstdefinestyle{mystyle}{
    backgroundcolor=\color{backcolour},   
    commentstyle=\color{codegreen},
    keywordstyle=\color{magenta},
    stringstyle=\color{codepurple},
    basicstyle=\ttfamily\footnotesize,
    breakatwhitespace=false,         
    breaklines=true,                 
    captionpos=b,                    
    keepspaces=true,                            showspaces=false,                
    showstringspaces=false,
    showtabs=false,                  
    tabsize=2
}

\usepackage{amsthm}

\theoremstyle{remark}

\usepackage[utf8]{inputenc}
\DeclareUnicodeCharacter{2460}{\textcircled{1}}
\DeclareUnicodeCharacter{2461}{\textcircled{2}}
\definecolor{Reasoning}{HTML}{8dcaf6}
\definecolor{Transition}{HTML}{f8827e}
\definecolor{Execution}{HTML}{fbc08a}

\DeclareUnicodeCharacter{221A}{\ensuremath{\sqrt{}}} 
\DeclareUnicodeCharacter{00B0}{\ensuremath{^\circ}} 
\DeclareUnicodeCharacter{2264}{\ensuremath{\leq}}   
\DeclareUnicodeCharacter{2265}{\ensuremath{\geq}}   
\DeclareUnicodeCharacter{2248}{\ensuremath{\approx}}
\DeclareUnicodeCharacter{2260}{\ensuremath{\neq}}   
\DeclareUnicodeCharacter{2019}{'}                  

\definecolor{violet9933FF}{HTML}{9933FF}
\definecolor{green00FF91}{HTML}{00FF91}
\definecolor{blue3333FF}{HTML}{3333FF}
\usepackage{tikz}

\newcommand{\bluecommenttext}[1]{\hfill\textcolor{blue}{$\blacktriangleright$~#1}}

\SetKwInput{KwInit}{Initialize}
\SetKwComment{tcp}{$\blacktriangleright$\ }{}   

\usepackage{hhline}

\newcommand*\circled[1]{\tikz[baseline=(char.base)]{
            \node[shape=circle,fill,inner sep=0pt] (char) {\textcolor{white}{#1}};}}
\usepackage[capitalize,noabbrev]{cleveref}
\crefformat{section}{\S#2#1#3} 
\crefformat{subsection}{\S#2#1#3}
\crefformat{subsubsection}{\S#2#1#3}

\title{Polestar: Drift-Aware Cache Calibration and Token Commitment for Efficient Inference of Diffusion LLMs}

\author{Mingyu Lee$^{*1}$, Akshat Ramachandran$^{*1}$, Souvik Kundu$^{2}$, \textbf{and} \textbf{Tushar Krishna}$^1$\\
$^{1}$Georgia Institute of Technology, USA \\
$^{2}$Intel AI Group, USA\\
$^{1}$\texttt{mlee864@gatech.edu, akshat.r@gatech.edu, tushar@ece.gatech.edu}\\
$^{2}$\texttt{souvikk.kundu@intel.com} \\
$^*$Equal Contribution}

\begin{document}

\maketitle

\begin{abstract}
The inference efficiency of diffusion large language models (dLLMs) is constrained by two challenges: bidirectional attention precludes efficient KV-cache reuse, while increasing decoding parallelism with static confidence thresholds can compromise generation quality. We observe that both challenges arise from a shared phenomenon: as tokens are decoded, their contextual integration through bidirectional attention causes token representations to drift (evolve) across decoding steps. This insight motivates \textbf{Polestar}, a training-free inference framework that uses \emph{token representation drift} as a unified signal to jointly address both challenges. Polestar comprises two components: \textbf{Polestar-Cache}, which identifies stale KV-cache positions via drift and performs sparse KV-cache refreshes to enable efficient reuse, and \textbf{Polestar-Commit}, which detects sharp drift events to reliably identify commit-ready tokens. Across mathematics and coding benchmarks on several dLLM families, Polestar sets a new state of the art on the accuracy--throughput Pareto frontier, achieving up to $\mathbf{10.73}\%$ accuracy improvement, up to $\mathbf{3.7}\times$ higher throughput, and high decoding parallelism of $3.67$ tokens per forward pass over existing baselines.
\end{abstract}

\section{Introduction}
\vspace{-2mm}
\label{sec:introduction}
\begin{wrapfigure}{R}{0.43\textwidth}
\vspace{-10mm}
  \centering \includegraphics[width = 0.43\textwidth, keepaspectratio]{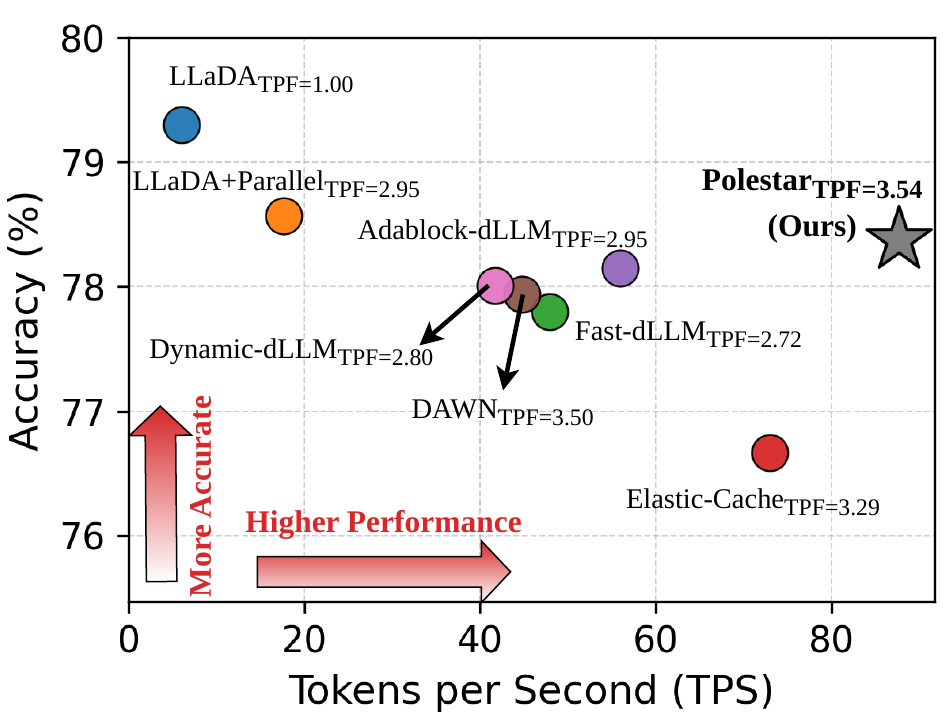}
  \vspace{-7mm}
  \caption{Accuracy-throughput (TPS) trade-off on GSM8K using LLaDA-8B-Instruct, with each method subscripted by its achieved tokens per forward pass (TPF).} 
  \vspace{-7mm}
  \label{fig:teaser}
\end{wrapfigure}

The inherent sequential dependency of autoregressive large language models (AR-LLMs) \cite{grattafiori2024llama, ramachandran2025microscopiq, tian2025skipkv} limits decoding parallelism and often leads to substantial inference performance degradation, particularly for long-form generation \cite{yuan2025native, ramachandran2025thinkv}. Recently, masked diffusion LLMs (dLLMs) \cite{nie2025large, ye2025dream} have emerged as an encouraging alternative to AR-LLMs. dLLMs enable parallel token decoding by reformulating generation as an iterative, multi-step denoising process \cite{austin2021structured} that progressively refines noisy tokens into clean predictions \cite{arriola2025block}.

However, dLLMs face a trade-off between parallelism and output quality, with the best accuracy often achieved when decoding one token per forward pass \cite{nie2025large, ye2025dream} (see LLaDA in \autoref{fig:teaser}). Moreover, their bidirectional attention precludes standard AR-LLM key–value (KV) caching,
necessitating KV recomputation at each decode step, resulting in reduced performance compared to AR-LLMs \cite{liu2025dllmcache}.

\begin{figure*}[!t]
  \centering \includegraphics[width = \linewidth, keepaspectratio]{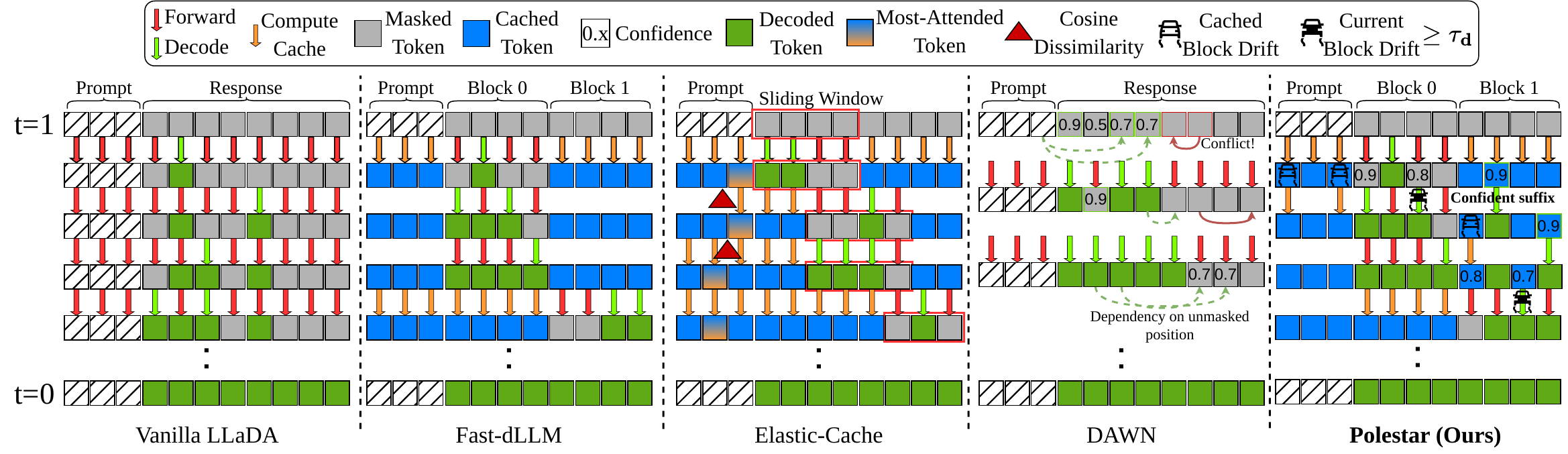}
  \vspace{-6mm}
  \caption{Illustration of dLLM inference strategies. \textit{Vanilla LLaDA} performs full-sequence recomputation at each step. \textit{Fast-dLLM} uses block-wise KV caching with confidence-based parallel decoding \cite{wu2025fast}. \textit{Elastic-Cache} triggers adaptive KV refreshes by monitoring increase in cosine dissimilarity of the most-attended tokens \cite{nguyen2025attention}. \textit{DAWN} uses dependency-aware decoding to improve decoding parallelism \cite{luo2026dawn}. \textbf{Polestar} employs token representation drift to selectively and sporadically refresh stale KV cache positions and commit tokens earlier in both the current and suffix block.}
  \vspace{-8mm}
  \label{fig:polestar_intro}
\end{figure*}

Existing techniques explore training-free inference optimizations to address the above inherent limitations of dLLMs along \underline{two primary axes}: \circled{A} reducing per-step latency, and \circled{B} increasing decoding parallelism to reduce the number of steps. \autoref{fig:polestar_intro} illustrates contemporary strategies across the two axes for optimizing dLLM inference. We defer an extended discussion to \Cref{appendix_related_work}

\noindent
\textit{Reducing per-step latency: }Prior works reduce the computational cost of a decoding step via KV cache reuse with selective updates to mitigate cache staleness. Fast-dLLM \cite{wu2025fast} adopts a block-wise caching that reuses KV representations outside the current decoding block with periodic refresh at block boundaries. Elastic-Cache \cite{nguyen2025attention} employs a sliding-window scheme with adaptive, layer-wise updates based on signals from the most-attended tokens. d$^2$Cache \cite{jiang2025d} selectively updates KV cache using token-level uncertainty. 
However, these methods either assume static representations within a block (\cite{wu2025fast}) or collect coarse proxy signals from individual tokens (\cite{nguyen2025attention}) or layer statistics (\cite{wudynamic}), treating them as representative of the entire sequence. This ignores the joint evolution of token representations across steps and layers, leading to miscalculated cache validity. Thus, stale KV cache degrades prediction confidence and limits parallelism while requiring redundant computation.

\noindent
\textit{Increasing decoding parallelism: }Fast-dLLM \cite{wu2025fast} uses static confidence thresholds for token commitment, but fixed thresholds fail to adapt across decoding steps and inputs. Other works \cite{lu2025adablock, shen2025improving} adapt block sizes based on confidence volatility, but introduces irregular execution and reduced hardware efficiency. DAWN \cite{luo2026dawn} uses attention-derived dependency graphs for conflict-free updates, but incurs per-step graph construction overhead. KLASS \cite{kim2025klass} leverages output stability for commitment, but often enforces aggressive, unconstrained parallel decoding leading to accuracy degradation. Overall, these approaches are either insufficiently adaptive or incur higher per-step latencies (\autoref{fig:teaser}) to improve decoding parallelism.

\textbf{Our Contributions.} Motivated by the limitations of existing techniques, we present \textbf{Polestar}, a `\textit{training-free}' dLLM optimization method to jointly address \underline{both axes} of dLLM inference optimization. We identify and leverage a principled signal: \textbf{token representational drift}. We show that drift is an inherent characteristic of dLLM inference (\cref{sec:motivation}, \Cref{sec:theory}) induced by bidirectional attention, rather than merely an error artifact of KV caching \cite{nguyen2025attention}. Specifically, Polestar uses drift to answer two key questions: 1) \underline{\emph{when and where}} should stale KV cache be updated to enable efficient cache reuse in dLLMs, and 2) \underline{\emph{when}} can tokens be reliably committed to increase decoding parallelism without compromising output quality. Polestar makes the following key contributions:


\begin{itemize}[align=right,labelsep=2pt,labelwidth=1em,leftmargin=0.5em,nosep]

\item \textbf{Polestar-Cache (\cref{sec:polestar_cache}): }We propose a novel KV-cache refresh scheme that uses token representation drift in each layer's input hidden states to identify stale KV cache positions and perform sparse, sporadic KV updates with minimal overhead.   
\item \textbf{Polestar-Commit (\cref{sec:polestar_commit}): }We observe that sharp token-drift spikes correlate with the onset of convergence toward a token's final output representation. We leverage this signal to commit tokens earlier and more reliably than confidence-only decoding, improving decoding parallelism.   
\item \textbf{Polestar System Optimization (\cref{sec:system}):} We enable high-throughput Polestar execution through CPU offloading and latency hiding, effectively overlapping Polestar-specific operations with the main dLLM forward pass. 
\end{itemize}

Through algorithm--system co-design, Polestar sets a new state of the art on the accuracy--throughput Pareto frontier for dLLM inference (\autoref{fig:teaser}). Across mathematics and coding benchmarks (\cref{sec:results}) spanning multiple dLLM families, Polestar achieves up to $\mathbf{10.73\%}$ \textbf{accuracy improvement} ($3.74\%$ on average), up to $\mathbf{3.7\times}$ \textbf{higher TPS} ($1.6\times$ on average), and decoding parallelism of $\mathbf{3.67}$ \textbf{tokens per forward (TPF)} over existing SoTA baselines.   




\section{Preliminaries: Inference with dLLMs}
\vspace{-2mm}
\label{sec:preliminaries}

Let $\mathcal{V}$ be the token vocabulary augmented with special masking token $\texttt{[MASK]}$. Given a prompt $\mathbf{p} = (p_0, \ldots, p_{L_p-1}) \in \mathcal{V}^{L_p}$, the model generates $L_G$ tokens by iterative denoising over $T$ reverse-indexed steps $t \in \{T, T-1, \ldots, 0\}$. The initial sequence is $\mathbf{y}^{(T)} = (\mathbf{p}, \mathbf{m}_{L_G})$, where $\mathbf{m}_{L_G} \triangleq (\texttt{[MASK]})^{L_G}$.

\noindent
\textbf{Denoising. }Let $\mathcal{J} \triangleq \{0, 1, \ldots, L_p + L_G - 1\}$ be the full token index set. At each step $t$, a mask predictor $m_\theta$ \cite{nie2025large} generates a tentative reconstruction $\hat{\mathbf{y}}^{(t)}$ by greedily selecting the most likely token to be committed at every position:
\vspace{-2mm}
\begin{equation}
\hat{y}^{(t)}_i = \texttt{argmax}_{v \in \mathcal{V} \setminus \texttt{[MASK]}} m_\theta(v \mid \mathbf{y}^{(t)}, i), \quad \forall i \in \mathcal{J}.
\end{equation}
\vspace{-6mm}

\noindent
\textbf{Semi-autoregressive decoding. } Here, we partition $L_G$ into $\mathcal{B}$ contiguous blocks (indexed by $b$) of fixed size $B$. Decoding proceeds sequentially across blocks with tokens decoded in parallel within a block. Let $\mathcal{J}_{{b}} \subseteq \mathcal{J}$ denote the index set of the tokens in the current decode block. At step $t$, the masked positions within the block are $\mathcal{M}^{(t)} \triangleq \{ i \in \mathcal{J}_{{b}} \mid y^{(t)}_i = \texttt{[MASK]} \}$.
The model’s certainty in its provisional prediction $\hat{y}^{(t)}_i$ at position $i$ is measured by a confidence score $c^{(t)}_i$ \cite{wu2025fast}. Based on this score the position to unmask is determined by a sampler set $\mathcal{S}^{(t)} \triangleq \{ i \in \mathcal{M}^{(t)} \mid c^{(t)}_i \ge \tau \}$, where $\tau$ is a predefined confidence threshold.
The sequence is then updated as,
\vspace{-5mm}

\begin{equation}
y^{(t-1)}_i =
\begin{cases}
\hat{y}^{(t)}_i, & i \in \mathcal{S}^{(t)}, \\
\texttt{[MASK]}, & i \in \mathcal{M}^{(t)} \setminus \mathcal{S}^{(t)}, \\
y^{(t)}_i, & i \notin \mathcal{J}_{{b}}
\end{cases}
\end{equation}
\vspace{-3mm}

This yields $N^{(t)} \triangleq |\{ i \in \mathcal{S}^{(t)} \mid \hat{y}^{(t)}_i \neq \texttt{[EOS]} \}|$ tokens decoded at step $t$\footnote{We term the average $N^{(t)}$ over all decoding steps as the tokens per forward (TPF).}. The denoise-decode cycle is repeated until all positions in $\mathcal{J}_{{b}}$ are unmasked, and decoding advances to the next block. 

\textbf{KV Caching in dLLMs. }For layer $\ell$ at step $t$, let $\mathcal{I} \subseteq \mathcal{J}$ index cached keys and values, denoted $\mathbf{K}_{\mathcal{I}}^{(t,\ell)} = \{\mathbf{K}^{(t,\ell)} _ {[i]}\}_{i \in \mathcal{I}}$ and $\mathbf{V}_{\mathcal{I}}^{(t,\ell)} = \{\mathbf{V}^{(t,\ell)} _ {[i]}\}_{i \in \mathcal{I}}$. Under block-wise KV caching, prior to decoding a block, the KV representations for all token positions outside the current block ($b$) are computed once at block entry ($t_{b}$) and cached across multiple decoding steps. The cached token positions are given by $\mathcal{I} \triangleq \mathcal{J} \setminus \mathcal{J}_{{b}}$. At each decoding step $t$, KV tensors for tokens in the active block $\mathcal{J}_{{b}}$ are computed from current input hidden states $\mathbf{H}^{(t,\ell)}_{\mathcal{J}_{{b}}}$ via linear projections \cite{nguyen2025attention}.  Queries from the active block attend jointly to cached $\mathbf{K,V}$ and active-block $K,V$ tensors\footnote{We refer to non-cached KV as tensors and denote them with non-bold notation.}: 
\vspace{-2mm}
\begin{equation}
\mathbf{A}_{\mathcal{J}_{{b}}}^{(t,\ell)}
=
\mathrm{Attn}\!\left(
{Q}_{\mathcal{J}_{{b}}}^{(t,\ell)},
\;
\mathbf{K}_{\mathcal{I}}^{(t_{b},\ell)} \cup {K}_{\mathcal{J}_{{b}}}^{(t,\ell)},
\;
\mathbf{V}_{\mathcal{I}}^{(t_{b},\ell)} \cup {V}_{\mathcal{J}_{{b}}}^{(t,\ell)}
\right)
\end{equation}
After $\mathcal{J}_{{b}}$ is decoded, the cache is refreshed: KV representations of the completed block are inserted, while those corresponding to the next active block are removed, giving $\mathcal{I} \leftarrow (\mathcal{I} \cup \mathcal{J}_{{b}})\setminus \mathcal{J}_{{b+1}}$. Polestar adopts semi-autoregressive decoding with KV caching as its baseline inference procedure.

\begin{figure}[!t]
  \centering
  \makebox[\textwidth][c]{
    \includegraphics[width=\linewidth, keepaspectratio]{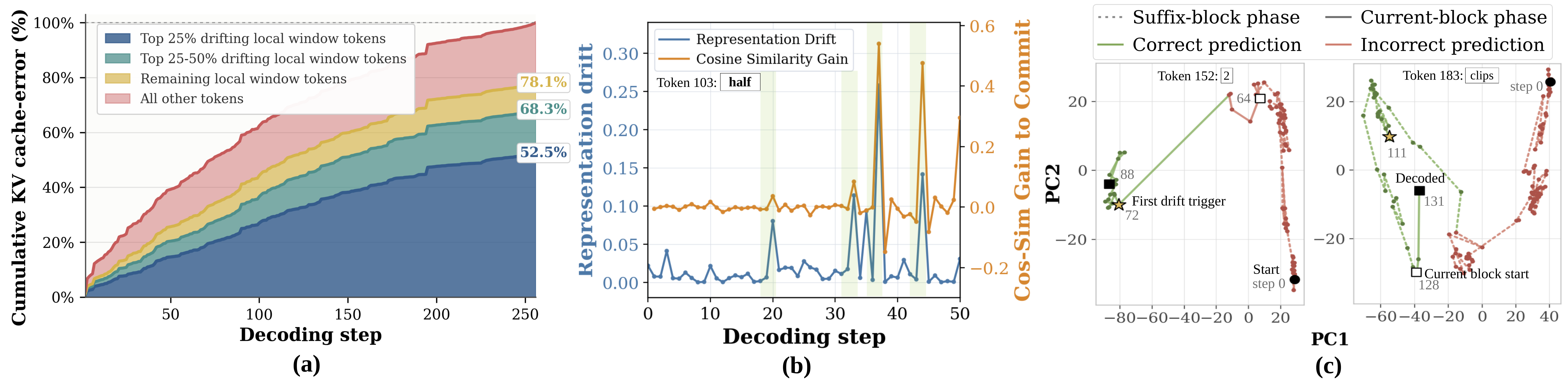}
  }
  \vspace{-5mm}
  \caption{Motivating analyses for Polestar: (a) KV-cache error concentrates on high-drift tokens near the current decoding block; (b) token representations advance toward their final commitment states through sharp drift events; and (c) drift events align with the onset of representational convergence.}
  \vspace{-6mm}
  \label{fig:motivation}
\end{figure}

\section{Motivational Analyses}
\vspace{-2mm}
\label{sec:motivation}

\textbf{Token Representation Drift.}In dLLMs, bidirectional attention makes token representations mutually dependent: a new token unmasking influences the context available to all other positions \cite{nie2025large} (\Cref{sec:theory}). Therefore, the internal representation (hidden states) of all other tokens are updated even when those tokens are not decoded at a particular step. We term this context-conditioned evolution as {\textit{token representation drift}}. 
%
Contemporary approaches \cite{nguyen2025attention,wudynamic} treat drift primarily as a KV-caching error artifact between cached and recomputed KV representations, and estimate it using cosine dissimilarity over a small token subset. 
However, it may produce a misleading signal when the chosen token subset remains stable while the broader context evolves, leading to accuracy degradation (\autoref{fig:teaser}).

Conversely, we quantify drift using KL-divergence, which directly captures how a token’s contextual distribution shifts over the full attention support (\cref{sec:ablations}, \Cref{sec:theory}). For token position $i$ at layer $\ell$ and step $t$, we measure drift as the KL-divergence of its attention distribution between two consecutive steps \footnote{Unless otherwise specified, all subsequent references to drift refer to KL-divergence-based token representation drift.}:
\vspace{-2mm}
\begin{equation}
\label{equation:drift}
{D}_{i,\ell}^{(t)}=\texttt{KL}\!\left(\mathbf{A}_{i,\ell}^{(t)} \,\|\, \mathbf{A}_{i,\ell}^{(t-1)}\right)
\end{equation}
\vspace{-6mm}

\subsection{Token Drift Reveals When and Where KV Cache Need Updating}
\vspace{-2mm}
\label{sec:drift_motivation}
\begin{wrapfigure}{R}{0.35\textwidth}
\vspace{-5mm}
  \centering
  \includegraphics[width=0.34\textwidth, keepaspectratio]{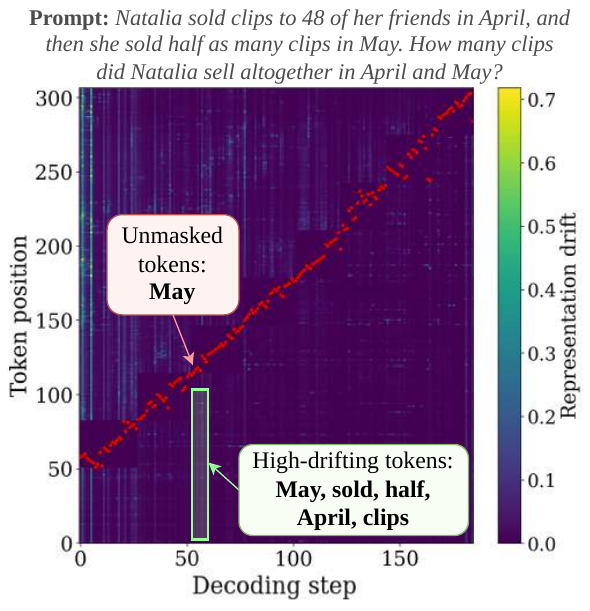}
  \vspace{-3mm}
  \caption{{Heatmap depicting drift across token positions and decoding steps, with unmasked tokens at every step shown in red.}}
  \vspace{-7mm}
  \label{fig:motivation_a}
\end{wrapfigure}
\textbf{Setup. } We use LLaDA-8B-Instruct \cite{nie2025large} with generation length of $256$ with $B = 32$ \cite{wu2025fast}, for input prompts from GSM8K \cite{cobbe2021training}. The baseline performs full-sequence forward passes with a TPF of 1 at each decoding step without cache reuse.

We first demonstrate that token drift, as quantified in \cref{equation:drift}, indeed captures contextual adaptation. \autoref{fig:motivation_a} presents drift as a heatmap for every token position across decoding steps for the given input prompt. Evident from \autoref{fig:motivation_a}, token representation drift is prominent in both the prefix and suffix (masked positions) region of the currently decoded token, demonstrating contextual integration. Notably, we observe higher drift at semantically related positions, which serves as a validation of the KL-div-based drift measurement. For instance, when token \textit{May} is unmasked, we observe higher drift in semantically related tokens (e.g., \textit{April}).

We next show how this token representation drift reveals KV-cache staleness for cache-enabled dLLM inference (\cref{sec:preliminaries}). For the purpose of this experiment, we measure staleness as the average cosine distance (equation in \Cref{sec:appendix_cosine_distance}) between cached KV and their corresponding recomputed KV tensors from full-sequence forward pass. \autoref{fig:motivation}(a) presents the cumulative KV-cache error across decoding steps, normalized to the error accumulated over the full generation. For this analysis, we group tokens by both location and drift magnitude. We define a local window around the current decoding block, consisting of the immediately preceding prefix block and the following suffix block. Within this window, we separate the top-$25\%$ and top-$25$--$50\%$ highest-drift tokens; the remaining two groups are the least drifting tokens in the local-window and all tokens outside the local window. \autoref{fig:motivation}(a) clearly shows that high-drift tokens near the current decoding block contribute disproportionately to the overall KV-cache error.
Building on this in \Cref{sec:appendix_selective_update}, we further show that local-window drift increases sharply when multiple tokens are decoded in a single forward pass, indicating that KV-cache updates are most needed after substantial contextual integration. \emph{\textbf{Polestar-Cache} leverages these findings to perform sparse and sporadic KV-cache updates for high-drift tokens in the vicinity of the current decoding block.}     

\subsection{Token Drift Identifies Opportunities for Early Commitment}
\vspace{-2mm}
\label{sec:commit_motivation}
Let $z_i^{(t)}$ represents the centered log-probabilities of the output logits for each token position $i$ at step $t$. $z_i^{(T_i)}$ denotes the predicted state once the token is committed. For the purpose of this analysis, we quantify commitment progress at step $t$ as the increase in cosine similarity to the final commitment state from step $t-1$ to $t$ as, $\Delta \mathcal{C}_i^{(t)} =
\texttt{cos\_sim}\!\left(z_i^{(t)}, z_i^{(T_i)}\right)
-
\texttt{cos\_sim}\!\left(z_i^{(t-1)}, z_i^{(T_i)}\right)$.

Following the same setup in \cref{sec:drift_motivation}, we measure the relationship between token representation drift and $\Delta \mathcal{C}_i^{(t)}$ for a particular token in \autoref{fig:motivation}(b). We observe that progress toward commitment is not gradual, but occurs through sharp drift events. In particular, a token moves substantially closer to its commitment state when its drift rises significantly above its recent drift history in the preceding steps.

We further examine how this reveals opportunities for early commitment of tokens. Following \cite{jiang2025d}, \autoref{fig:motivation}(c) visualizes PCA trajectories of last-layer value states across decoding steps for two token positions. We observe that the first sharp drift trigger (yellow star) occurs at the onset of representational convergence, when the token trajectory begins to stabilize toward its decoded state. Notably, for token 183 in \autoref{fig:motivation}(c), this occurs while the token is still in the suffix region. Overall, the trajectories show that drift-based commitment enables substantially earlier token commitment than static confidence-based commitment with threshold $\tau_s = 0.9$ (black square) \cite{wu2025fast}. \emph{\textbf{Polestar-Commit} exploits this behavior to identify tokens for early commitment both in current and suffix block.}

\begin{figure}[t]
  \centering
  \makebox[\textwidth][c]{
    \includegraphics[width=0.98\linewidth, keepaspectratio]{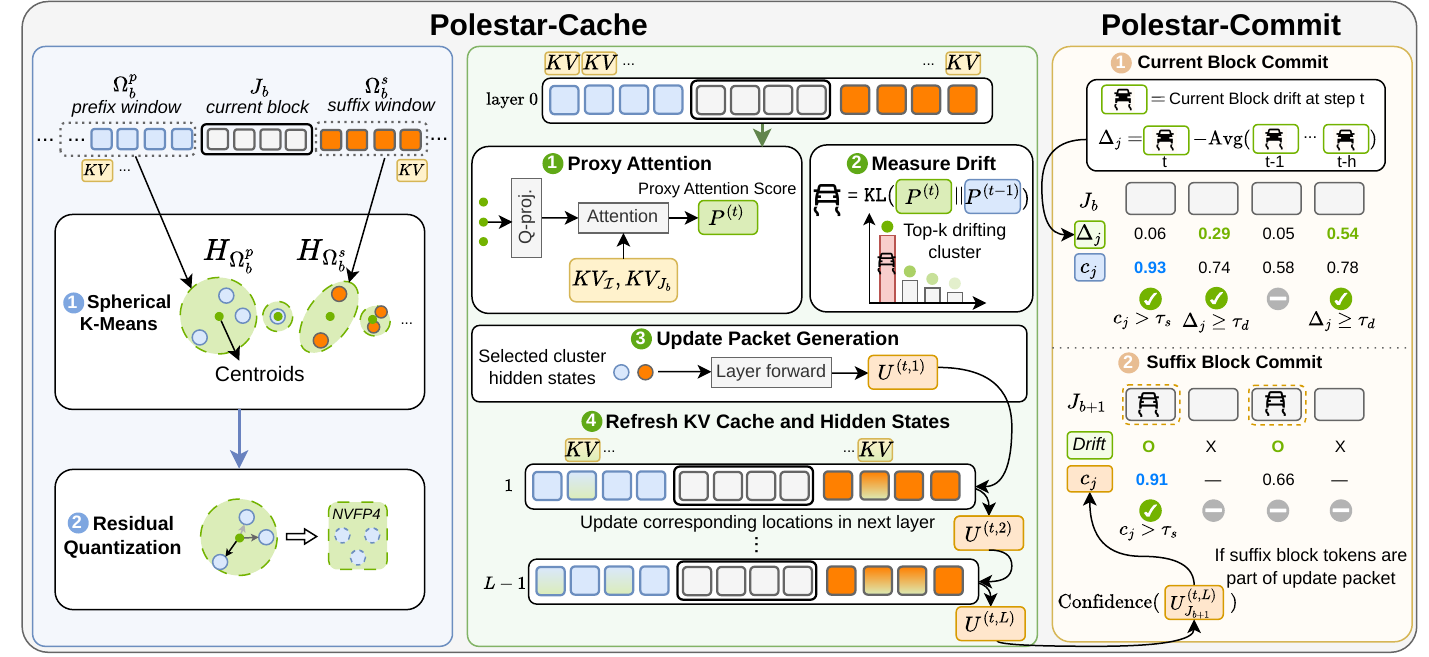}
  }
  \vspace{-4.7mm}
  \caption{Overview of Polestar. System optimizations omitted for brevity.}
  \vspace{-6mm}
  \label{fig:methodology}
\end{figure}

\section{Polestar Methodology}
\vspace{-2mm}
This section presents the Polestar methodology, including its algorithmic design (\textbf{Polestar-Cache} and \textbf{Polestar-Commit}) and system optimizations. \autoref{fig:methodology} illustrates the overall framework, and \Cref{alg:polestar} summarizes the methodology.
\vspace{-3mm}
\subsection{Polestar-Cache}
\vspace{-2mm}
\label{sec:polestar_cache}

\noindent
\textbf{Initialization. }At the start of the decoding phase, we perform a full-sequence forward pass to compute and cache the KV representations for all layers, of all token positions outside the current block $\mathcal{J}_b$. Keys and values of the current block are not cached and computed at every step.

\noindent
\textbf{Block entry. }As observed in \cref{sec:drift_motivation}, majority of the KV-cache error is concentrated within a local window around the currently decoding block. Therefore, we maintain a local window $\Omega_b = \Omega^p_b\cup\Omega^s_b$, composed of prefix and suffix windows as shown in \autoref{fig:methodology}. At the entry of each block, we cache the input hidden states $\mathbf{H}_{\Omega_b}$ of all layers for all token positions within the local window. Due to residual connections in the dLLM forward pass \cite{nie2025large}, we cache input hidden states in addition to KV cache to ensure that layer outputs are updated consistently with the evolving contextual representations.    

\noindent
\textbf{Cluster and Quantize Hidden States. }During decoding, measuring drift for all tokens in the local window introduces significant overheads. To mitigate this, we use spherical K-means \cite{ramachandran2025thinkv,hooper2025multipole} to cluster $\mathbf{H}_{\Omega_b}$ in the local window. We then employ the computed cluster centroids as sufficient and compact representations (proxies) of the local window tokens for measuring drift. Furthermore, to minimize the memory overhead of caching the hidden states, for each cluster, we quantize the residuals between its hidden states and the corresponding centroid. Each residual is quantized along the channel dimension using NVFP4 \cite{ramachandran2025thinkv} (\Cref{sec:appendix_quantization}).

\noindent
\textbf{Measuring Drift and Layer-wise KV Updates. }At each step $t$ and layer $\ell$, we project all cluster centroids into query tokens and compute the ``proxy'' attention distribution $P^{(t,\ell)}$ using a combination of cached KV and recomputed KV tensors for the active block. Following \cref{sec:drift_motivation}, we select the top-$k$ clusters by drift magnitude to identify token positions requiring updates. We then perform a sparse forward pass only over the hidden states associated with these clusters to obtain updated layer $\ell$ outputs \footnote{This sparse forward pass is necessary because, under semi-autoregressive decoding, the standard forward pass only processes tokens in the active block (\cref{sec:preliminaries}).}. As shown in \autoref{fig:methodology}, this is layer $\ell$'s update packet ${U^{(t,\ell+1)}}$. Layer $\ell+1$ uses this update packet to refresh corresponding stale cached hidden states and subsequently update stale KV-cache entries within the local window. The corresponding centroids are also updated incrementally to reflect the refreshed hidden states. Repeating this process across layers yields sparse layer-wise KV-cache updates that keep the cache aligned with the evolving context without full-sequence recomputation.

\noindent
\textbf{Update Schedule. }Unlike existing techniques \cite{wu2025fast, nguyen2025attention} that refresh the KV cache at fixed intervals, Polestar-Cache uses token drift to selectively refresh only when cached states are likely to become stale (\Cref{sec:appendix_selective_update}). Polestar-Cache refreshes the cache when newly decoded context is expected to cause substantial misalignment: either after a step that decodes more than $\tau_{\mathrm{upd}}$ tokens, or when the cumulative number of decoded tokens since the last refresh exceeds $\tau_{\mathrm{upd}}$. Additionally, since tokens outside the local window still contribute non-negligible KV-cache error (\autoref{fig:motivation}(a)), Polestar-Cache performs a full-sequence forward pass at every alternate block entry to refresh non-local positions.

\vspace{-2mm}
\subsection{Polestar-Commit}
\vspace{-2mm}
\label{sec:polestar_commit}

\noindent
\textbf{Current Block Commit. }Existing techniques \cite{wu2025fast, nguyen2025attention} commit current-block tokens using a static confidence threshold $\tau_s=0.9$. In contrast, Polestar-Commit performs drift-aware token commitment using final-layer's token drift score, which provides a signal closer to the model's eventual prediction. To detect \underline{sharp} drift events, we maintain the mean drift over the past $h$ steps and define the drift delta $\Delta_j^{(t)}$ as the deviation of the current drift from this historical mean. A positive $\Delta_j^{(t)}$ indicates that masked position $j$ is undergoing a stronger contextual update than its recent history. We then modulate this signal using a confidence-conditioned dynamic threshold, $\tau_{d,j}^{(t)} = \alpha(\tau_s - c_j^{(t)})^2$, where $\alpha$ is a scaling factor and $c_j^{(t)}$ is the prediction confidence for position $j$. A current-block token is committed when $\Delta_j^{(t)} \geq \tau_{d,j}^{(t)}$. Thus, tokens with confidence at least $\tau_s$ are committed as in prior methods \cite{wu2025fast} since the dynamic threshold tends to 0. More importantly, lower-confidence tokens can still be committed when they exhibit a sufficiently strong drift event, indicating representational convergence that confidence alone fails to capture. Tokens farther from the static threshold require stronger drift events for commitment and vice versa.

\noindent
\textbf{Suffix Block Commit. }In \cref{sec:polestar_cache}, each layer generates an update packet. The final-layer update packet is not propagated further, since there is no subsequent layer to update. However, we observe that it provides a useful signal for suffix-block commitment: when suffix-window centroids are selected among the top-$k$ drifting clusters, tokens in the suffix region are undergoing strong contextual updates. For each identified high-drift suffix position, we obtain logits from the final-layer update packet and evaluate its confidence. Since suffix tokens are not actively denoised at the current step, their high-drift events primarily arise from contextual integration with semantically related tokens decoded in the current block. We therefore adopt a conservative commitment rule: a suffix token $j$ is committed only if $c_j^{(t)} \geq \tau_s$. Thus, drift determines \emph{which} suffix positions become eligible for commitment, while confidence determines \emph{which} eligible positions are safe to commit.

\vspace{-2mm}
\subsection{Polestar System Optimization}
\vspace{-2mm}
\label{sec:system}

\noindent
\textbf{CPU Offloading. }As described in \cref{sec:polestar_cache}, Polestar-Cache uses cluster centroids to identify the top-$k$ drifting clusters to selectively refresh corresponding KV-cache entries. Because the cached hidden states introduce an additional memory footprint that competes for memory with the KV cache on GPU, Polestar stores only the centroids on GPU and offloads the quantized hidden-state clusters to pinned CPU memory. During decoding, only the top-$k$ drifting clusters are fetched asynchronously from CPU memory. Since the hidden-state clusters are quantized, it significantly reduces CPU-GPU transfer bandwidth. Following \cite{ramachandran2025thinkv}, the retrieved hidden states' dequantization is fused with update packet computation to reduce overhead.

\noindent
\textbf{Latency Hiding. }To reduce Polestar's runtime overhead, we leverage CUDA streams to overlap additional  operations with the main dLLM forward pass. The main stream executes the current block attention and FFN computation. Since Polestar performs different operations at block entry and during subsequent within-block steps, we optimize the two cases separately. At block entry, auxiliary streams perform hidden-state caching, clustering, quantization, and offloading in parallel with the main attention computation. During within-block steps, proxy-attention computation, drift measurement, and update-packet generation are similarly executed on auxiliary streams. Since the token positions selected in the current layer and those that are refreshed via the update packet in the subsequent layer are identical, Polestar prefetches the corresponding clusters for both the current and next layers from CPU memory, allowing subsequent-layer refresh to proceed without stalling. \autoref{fig:performance_breakdown}(b,c) in \Cref{sec:appendix_performance_breakdown} illustrates Polestar's optimized execution flow.
\begin{table*}[t]
\centering
\caption{Accuracy (\%), tokens-per-forward (TPF) and tokens-per-second (TPS) comparison of Polestar, with SoTA dLLM KV caching baselines. Speedup over baseline is indicated as \textcolor[HTML]{D35400}{($\cdot \times$)}.}
\renewcommand*{\arraystretch}{1.1}
\setlength{\tabcolsep}{4pt}
\resizebox{0.75\textwidth}{!}{%
\begin{tabular}{lc l|ccc|ccc}
\Xhline{2\arrayrulewidth}
\multirow{2}{*}{Benchmark} & \multirow{2}{*}{Gen Len} & \multirow{2}{*}{Method}
& \multicolumn{3}{c|}{LLaDA-8B-Instruct}
& \multicolumn{3}{c}{Dream-7B-Instruct} \\
& &
& Acc(\%)$\uparrow$ & TPF$\uparrow$ & TPS$\uparrow$
& Acc(\%)$\uparrow$ & TPF$\uparrow$ & TPS$\uparrow$ \\
\Xhline{2\arrayrulewidth}

\multirow{12}{*}{\makecell{GSM8K \\ \textit{(5-shot)}}}
& \multirow{6}{*}{256}
& \cellcolor[HTML]{D3D3D3}Baseline
& \cellcolor[HTML]{D3D3D3}79.30 & \cellcolor[HTML]{D3D3D3}1.00 & \cellcolor[HTML]{D3D3D3}6.58 \textcolor[HTML]{D35400}{(1.0$\times$)}

& \cellcolor[HTML]{D3D3D3}75.85 & \cellcolor[HTML]{D3D3D3}1.00 & \cellcolor[HTML]{D3D3D3}9.85 \textcolor[HTML]{D35400}{(1.0$\times$)}\\

& & \cellcolor[HTML]{D3D3D3}Baseline+Parallel
& \cellcolor[HTML]{D3D3D3}78.57 & \cellcolor[HTML]{D3D3D3}2.95 & \cellcolor[HTML]{D3D3D3}17.65 \textcolor[HTML]{D35400}{(2.7$\times$)}
& \cellcolor[HTML]{D3D3D3}72.96 & \cellcolor[HTML]{D3D3D3}1.56 & \cellcolor[HTML]{D3D3D3}12.10 \textcolor[HTML]{D35400}{(1.2$\times$)} \\
& & Fast-dLLM
& 77.88 & 2.72 & 47.88 \textcolor[HTML]{D35400}{(7.3$\times$)}
& 66.67 & 1.71 & 39.86 \textcolor[HTML]{D35400}{(4.0$\times$)} \\
& & Dynamic-dLLM
& 78.01 & 2.79 & 49.15 \textcolor[HTML]{D35400}{(7.5$\times$)}
& 68.16 & 1.79 & 41.01 \textcolor[HTML]{D35400}{(4.1$\times$)} \\
& & Elastic-Cache
& 77.58 & 2.93 & 55.01 \textcolor[HTML]{D35400}{(8.4$\times$)}
& 63.63 & 1.48 & 31.50 \textcolor[HTML]{D35400}{(3.1$\times$)} \\
& & \cellcolor[HTML]{D5E8D4}\textbf{Polestar (Ours)}
& \cellcolor[HTML]{D5E8D4}\textbf{78.33} & \cellcolor[HTML]{D5E8D4}\textbf{3.67} & \cellcolor[HTML]{D5E8D4}\textbf{87.57} \textcolor[HTML]{D35400}{\textbf{(13.3$\times$)}}
& \cellcolor[HTML]{D5E8D4}\textbf{72.40} & \cellcolor[HTML]{D5E8D4}\textbf{2.39} & \cellcolor[HTML]{D5E8D4}\textbf{52.80} \textcolor[HTML]{D35400}{\textbf{(5.4$\times$)}} \\
\hhline{~--------}
& \multirow{6}{*}{512}
& \cellcolor[HTML]{D3D3D3}Baseline
& \cellcolor[HTML]{D3D3D3}77.50 & \cellcolor[HTML]{D3D3D3}1.00 & \cellcolor[HTML]{D3D3D3}3.93 \textcolor[HTML]{D35400}{(1.0$\times$)}
& \cellcolor[HTML]{D3D3D3}76.90 & \cellcolor[HTML]{D3D3D3}1.00 & \cellcolor[HTML]{D3D3D3}6.59 \textcolor[HTML]{D35400}{(1.0$\times$)} \\
& & \cellcolor[HTML]{D3D3D3}Baseline+Parallel
& \cellcolor[HTML]{D3D3D3}77.50 & \cellcolor[HTML]{D3D3D3}2.85 & \cellcolor[HTML]{D3D3D3}18.82 \textcolor[HTML]{D35400}{(4.8$\times$)}
& \cellcolor[HTML]{D3D3D3}69.95 & \cellcolor[HTML]{D3D3D3}1.96 & \cellcolor[HTML]{D3D3D3}15.99 \textcolor[HTML]{D35400}{(2.4$\times$)} \\
& & Fast-dLLM
& 76.52 & 2.58 & 38.07 \textcolor[HTML]{D35400}{(9.7$\times$)}
& 64.55 & 1.38 & 25.11 \textcolor[HTML]{D35400}{(3.8$\times$)} \\
& & Dynamic-dLLM
& 76.65 & 2.90 & 42.79 \textcolor[HTML]{D35400}{(10.9$\times$)}
& 65.03 & 1.47 & 26.20 \textcolor[HTML]{D35400}{(4.0$\times$)} \\
& & Elastic-Cache
& 76.97 & 2.61 & 38.44 \textcolor[HTML]{D35400}{(9.8$\times$)}
& 58.94 & 1.13 & 19.39 \textcolor[HTML]{D35400}{(2.9$\times$)} \\
& & \cellcolor[HTML]{D5E8D4}\textbf{Polestar (Ours)}
& \cellcolor[HTML]{D5E8D4}\textbf{78.18} & \cellcolor[HTML]{D5E8D4}\textbf{3.59} & \cellcolor[HTML]{D5E8D4}\textbf{80.60} \textcolor[HTML]{D35400}{\textbf{(20.5$\times$)}}
& \cellcolor[HTML]{D5E8D4}\textbf{69.65} & \cellcolor[HTML]{D5E8D4}\textbf{2.13} & \cellcolor[HTML]{D5E8D4}\textbf{34.44} \textcolor[HTML]{D35400}{\textbf{(5.2$\times$)}} \\
\Xhline{1\arrayrulewidth}

\multirow{12}{*}{\makecell{MATH \\ \textit{(4-shot)}}}
& \multirow{6}{*}{256}
& \cellcolor[HTML]{D3D3D3}Baseline
& \cellcolor[HTML]{D3D3D3}33.50 & \cellcolor[HTML]{D3D3D3}1.00 & \cellcolor[HTML]{D3D3D3}9.10 \textcolor[HTML]{D35400}{(1.0$\times$)}
& \cellcolor[HTML]{D3D3D3}41.69 & \cellcolor[HTML]{D3D3D3}1.00 & \cellcolor[HTML]{D3D3D3}10.65 \textcolor[HTML]{D35400}{\textbf{(1.0$\times$)}} \\
& & \cellcolor[HTML]{D3D3D3}Baseline+Parallel
& \cellcolor[HTML]{D3D3D3}33.10 & \cellcolor[HTML]{D3D3D3}2.51 & \cellcolor[HTML]{D3D3D3}22.90 \textcolor[HTML]{D35400}{(2.5$\times$)}
& \cellcolor[HTML]{D3D3D3}41.30 & \cellcolor[HTML]{D3D3D3}1.68 & \cellcolor[HTML]{D3D3D3}25.87 \textcolor[HTML]{D35400}{(2.4$\times$)} \\
& & Fast-dLLM
& 30.18 & 2.30 & 38.13 \textcolor[HTML]{D35400}{(4.2$\times$)}
& 37.19 & 1.79 & 49.08 \textcolor[HTML]{D35400}{(4.6$\times$)} \\
& & Dynamic-dLLM
& 30.09 & 2.60 & 43.24 \textcolor[HTML]{D35400}{(4.8$\times$)}
& 37.47 & 2.04 & 54.93 \textcolor[HTML]{D35400}{(5.2$\times$)} \\
& & Elastic-Cache
& 29.35 & 2.08 & 42.56 \textcolor[HTML]{D35400}{(4.7$\times$)}
& 30.02 & 1.73 & 43.05 \textcolor[HTML]{D35400}{(4.0$\times$)} \\
& & \cellcolor[HTML]{D5E8D4}\textbf{Polestar (Ours)}
& \cellcolor[HTML]{D5E8D4}\textbf{32.77} & \cellcolor[HTML]{D5E8D4}\textbf{2.91} & \cellcolor[HTML]{D5E8D4}\textbf{70.05} \textcolor[HTML]{D35400}{\textbf{(7.7$\times$)}}
& \cellcolor[HTML]{D5E8D4}\textbf{40.75} & \cellcolor[HTML]{D5E8D4}\textbf{2.29} & \cellcolor[HTML]{D5E8D4}\textbf{59.62} \textcolor[HTML]{D35400}{\textbf{(5.6$\times$)}} \\
\hhline{~--------}
& \multirow{6}{*}{512}
& \cellcolor[HTML]{D3D3D3}Baseline
& \cellcolor[HTML]{D3D3D3}36.60 & \cellcolor[HTML]{D3D3D3}1.00 & \cellcolor[HTML]{D3D3D3}8.30 \textcolor[HTML]{D35400}{(1.0$\times$)}
& \cellcolor[HTML]{D3D3D3}38.95 & \cellcolor[HTML]{D3D3D3}1.00 & \cellcolor[HTML]{D3D3D3}9.80 \textcolor[HTML]{D35400}{\textbf{(1.0$\times$)}} \\
& & \cellcolor[HTML]{D3D3D3}Baseline+Parallel
& \cellcolor[HTML]{D3D3D3}36.15 & \cellcolor[HTML]{D3D3D3}2.50 & \cellcolor[HTML]{D3D3D3}19.60 \textcolor[HTML]{D35400}{(2.4$\times$)}
& \cellcolor[HTML]{D3D3D3}37.90 & \cellcolor[HTML]{D3D3D3}2.10 & \cellcolor[HTML]{D3D3D3}32.80 \textcolor[HTML]{D35400}{(3.3$\times$)} \\
& & Fast-dLLM
& 33.20 & 2.60 & 50.79 \textcolor[HTML]{D35400}{(6.1$\times$)}
& 32.64 & 2.27 & 57.79 \textcolor[HTML]{D35400}{(5.9$\times$)} \\
& & Dynamic-dLLM
& 32.22 & 2.79 & 54.57 \textcolor[HTML]{D35400}{(6.6$\times$)}
& 33.88 & 2.47 & 61.58 \textcolor[HTML]{D35400}{(6.3$\times$)} \\
& & Elastic-Cache
& 30.85 & 2.17 & 39.22 \textcolor[HTML]{D35400}{(4.7$\times$)}
& 33.17 & 2.35 & 54.46 \textcolor[HTML]{D35400}{(5.6$\times$)} \\
& & \cellcolor[HTML]{D5E8D4}\textbf{Polestar (Ours)}
& \cellcolor[HTML]{D5E8D4}\textbf{34.85} & \cellcolor[HTML]{D5E8D4}\textbf{3.30} & \cellcolor[HTML]{D5E8D4}\textbf{69.84} \textcolor[HTML]{D35400}{\textbf{(8.4$\times$)}}
& \cellcolor[HTML]{D5E8D4}\textbf{35.71} & \cellcolor[HTML]{D5E8D4}\textbf{2.81} & \cellcolor[HTML]{D5E8D4}\textbf{66.21} \textcolor[HTML]{D35400}{\textbf{(6.8$\times$)}} \\
\Xhline{1\arrayrulewidth}

\multirow{12}{*}{\makecell{HumanEval \\ \textit{(0-shot)}}}
& \multirow{6}{*}{256}
& \cellcolor[HTML]{D3D3D3}Baseline
& \cellcolor[HTML]{D3D3D3}43.35 & \cellcolor[HTML]{D3D3D3}1.00 & \cellcolor[HTML]{D3D3D3}17.60 \textcolor[HTML]{D35400}{(1.0$\times$)}
& \cellcolor[HTML]{D3D3D3}58.80 & \cellcolor[HTML]{D3D3D3}1.00 & \cellcolor[HTML]{D3D3D3}23.30 \textcolor[HTML]{D35400}{\textbf{(1.0$\times$)}} \\
& & \cellcolor[HTML]{D3D3D3}Baseline+Parallel
& \cellcolor[HTML]{D3D3D3}43.90 & \cellcolor[HTML]{D3D3D3}3.11 & \cellcolor[HTML]{D3D3D3}56.41 \textcolor[HTML]{D35400}{(3.2$\times$)}
& \cellcolor[HTML]{D3D3D3}58.80 & \cellcolor[HTML]{D3D3D3}2.10 & \cellcolor[HTML]{D3D3D3}46.97 \textcolor[HTML]{D35400}{(2.0$\times$)} \\
& & Fast-dLLM
& 38.63 & 2.64 & 54.19 \textcolor[HTML]{D35400}{(3.1$\times$)}
& 48.48 & 1.36 & 40.45 \textcolor[HTML]{D35400}{(1.7$\times$)} \\
& & Dynamic-dLLM
& 38.84 & 3.11 & 63.79 \textcolor[HTML]{D35400}{(3.6$\times$)}
& 48.00 & 1.45 & 42.21 \textcolor[HTML]{D35400}{(1.8$\times$)} \\
& & Elastic-Cache
& 41.67 & 3.08 & 72.75 \textcolor[HTML]{D35400}{(4.1$\times$)}
& 55.30 & 1.27 & 33.58 \textcolor[HTML]{D35400}{(1.4$\times$)} \\
& & \cellcolor[HTML]{D5E8D4}\textbf{Polestar (Ours)}
& \cellcolor[HTML]{D5E8D4}\textbf{42.42} & \cellcolor[HTML]{D5E8D4}\textbf{3.38} & \cellcolor[HTML]{D5E8D4}\textbf{87.45} \textcolor[HTML]{D35400}{\textbf{(5.0$\times$)}}
& \cellcolor[HTML]{D5E8D4}\textbf{57.69} & \cellcolor[HTML]{D5E8D4}\textbf{1.79} & \cellcolor[HTML]{D5E8D4}\textbf{55.18} \textcolor[HTML]{D35400}{\textbf{(2.4$\times$)}} \\
\hhline{~--------}
& \multirow{6}{*}{512}
& \cellcolor[HTML]{D3D3D3}Baseline
& \cellcolor[HTML]{D3D3D3}44.10 & \cellcolor[HTML]{D3D3D3}1.00 & \cellcolor[HTML]{D3D3D3}8.57 \textcolor[HTML]{D35400}{(1.0$\times$)}
& \cellcolor[HTML]{D3D3D3}58.80 & \cellcolor[HTML]{D3D3D3}1.00 & \cellcolor[HTML]{D3D3D3}15.48 \textcolor[HTML]{D35400}{\textbf{(1.0$\times$)}} \\
& & \cellcolor[HTML]{D3D3D3}Baseline+Parallel
& \cellcolor[HTML]{D3D3D3}44.10 & \cellcolor[HTML]{D3D3D3}2.71 & \cellcolor[HTML]{D3D3D3}34.53 \textcolor[HTML]{D35400}{(4.0$\times$)}
& \cellcolor[HTML]{D3D3D3}56.95 & \cellcolor[HTML]{D3D3D3}1.93 & \cellcolor[HTML]{D3D3D3}26.77 \textcolor[HTML]{D35400}{(1.7$\times$)} \\
& & Fast-dLLM
& 44.24 & 2.59 & 58.87 \textcolor[HTML]{D35400}{(6.9$\times$)}
& 51.51 & 1.23 & 32.21 \textcolor[HTML]{D35400}{(2.1$\times$)} \\
& & Dynamic-dLLM
& 44.48 & 2.49 & 56.74 \textcolor[HTML]{D35400}{(6.6$\times$)}
& 51.45 & 1.42 & 36.55 \textcolor[HTML]{D35400}{(2.4$\times$)} \\
& & Elastic-Cache
& 45.45 & 2.74 & 59.36 \textcolor[HTML]{D35400}{(6.9$\times$)}
& 53.03 & 1.01 & 23.14 \textcolor[HTML]{D35400}{(1.5$\times$)} \\
& & \cellcolor[HTML]{D5E8D4}\textbf{Polestar (Ours)}
& \cellcolor[HTML]{D5E8D4}\textbf{47.73} & \cellcolor[HTML]{D5E8D4}\textbf{3.23} & \cellcolor[HTML]{D5E8D4}\textbf{77.67} \textcolor[HTML]{D35400}{\textbf{(9.1$\times$)}}
& \cellcolor[HTML]{D5E8D4}\textbf{54.36} & \cellcolor[HTML]{D5E8D4}\textbf{1.77} & \cellcolor[HTML]{D5E8D4}\textbf{41.68} \textcolor[HTML]{D35400}{\textbf{(2.7$\times$)}} \\

\Xhline{2\arrayrulewidth}
\end{tabular}}
\label{tab:final_results}
\vspace{-5mm}
\end{table*}

\section{Experimental Evaluations}
\vspace{-2mm}
\label{sec:results}
\subsection{Experimental Setup}
\vspace{-2mm}
\noindent
\textbf{Models and Datasets. }We evaluate Polestar on LLaDA-8B-Instruct \cite{nie2025large}, LLaDA-1.5 \cite{zhu2025llada15}, Dream-7B-Instruct \cite{ye2025dream}, and multimodal LLaDA-V \cite{you2025lladav}. Text-only evaluation covers 5-shot GSM8K \cite{cobbe2021training}, 4-shot MATH \cite{hendrycks2021math}, ParallelBench \cite{kang2025parallelbench}, 0-shot HumanEval (pass@1) \cite{chen2021humaneval}, and 3-shot MBPP (pass@1) \cite{austin2021mbpp}. For LLaDA-V, we evaluate on MathVista \cite{lu2024mathvista} and MathVerse \cite{zhang2024mathverse}.

\noindent
\textbf{Hyperparameters. }We set block size $B=32$, static confidence threshold $\tau_s=0.9$ \cite{wu2025fast}, and local window $|\Omega_b|=3B$, with prefix window $|\Omega^p_b|=2B$ and suffix window $|\Omega^s_b|=B$. For $K$-means clustering, we use $K=8$ and select top-$k=4$ drifting clusters. We set the update threshold to $\tau_{\mathrm{upd}}=3$ tokens. For current-block commitment, we use drift history length $h=5$ and dynamic-threshold scaling factor $\alpha=10$ to bring confidence values to the same numeric scale as drift.

\noindent
\textbf{Baselines. }We compare Polestar against methods targeting per-step latency reduction via KV caching and methods targeting increased decoding parallelism. KV-caching baselines include Fast-dLLM~\cite{wu2025fast}, Dynamic-dLLM~\cite{wudynamic}, Elastic-Cache~\cite{nguyen2025attention}, d$^2$Cache~\cite{jiang2025d}, and EntropyCache~\cite{cheong2026entropycache}. Parallel-decoding baselines include baseline model with confidence-based parallel decoding, Fast-dLLM~\cite{wu2025fast}, AdaBlock-dLLM~\cite{lu2025adablock}, DAWN~\cite{luo2026dawn}, and KLASS~\cite{kim2025klass}.


\noindent
\textbf{Evaluation Setup. }All experiments use a single NVIDIA A100 80GB GPU or an NVIDIA GH200 Superchip. Unless otherwise specified, throughput measurements are reported on A100; please see \Cref{sec:appendix_gh200} for GH200. Following \cite{wu2025fast, nguyen2025attention}, we evaluate generation lengths of 256 and 512 with batch size 1. All models and methods are evaluated using lm-eval \cite{biderman2024lmeval} for consistency (see \Cref{sec:appendix_experiments} for details). 

\subsection{Main Results}

\begin{wraptable}{R}{0.5\textwidth}
\centering
\vspace{-5.5mm}
\caption{Polestar compared against SoTA dLLM parallel decoding baselines on LLaDA-1.5.}
\renewcommand*{\arraystretch}{1.1}
\setlength{\tabcolsep}{4pt}
\resizebox{\linewidth}{!}{%
\begin{tabular}{l c l|ccc}
\Xhline{2\arrayrulewidth}
{Benchmark} & {Gen Len} & {Method} &Acc(\%)$\uparrow$ & TPF$\uparrow$ & TPS$\uparrow$ \\
\Xhline{2\arrayrulewidth}

\multirow{5}{*}{\makecell{GSM8K \\ \textit{(5-shot)}}}
& \multirow{5}{*}{256}
& \cellcolor[HTML]{D3D3D3}Baseline
& \cellcolor[HTML]{D3D3D3}81.36 & \cellcolor[HTML]{D3D3D3}1.00 & \cellcolor[HTML]{D3D3D3}5.58 \textcolor[HTML]{D35400}{(1.0$\times$)} \\
& & \cellcolor[HTML]{D3D3D3}Baseline+Parallel
& \cellcolor[HTML]{D3D3D3}80.95 & \cellcolor[HTML]{D3D3D3}1.95 & \cellcolor[HTML]{D3D3D3}32.49 \textcolor[HTML]{D35400}{(5.8$\times$)} \\
& & DAWN
& 80.82 & 2.10 & 43.24 \textcolor[HTML]{D35400}{(7.7$\times$)} \\
& & KLASS
& 77.63 & 1.58 & 22.63 \textcolor[HTML]{D35400}{(4.1$\times$)} \\
& & \cellcolor[HTML]{D5E8D4}\textbf{Polestar (Ours)}
& \cellcolor[HTML]{D5E8D4}\textbf{81.06} & \cellcolor[HTML]{D5E8D4}\textbf{3.40} & \cellcolor[HTML]{D5E8D4}\textbf{79.65} \textcolor[HTML]{D35400}{\textbf{(14.3$\times$)}} \\

\Xhline{1\arrayrulewidth}

\multirow{5}{*}{\makecell{MBPP \\ \textit{(3-shot)}}}
& \multirow{5}{*}{256}
& \cellcolor[HTML]{D3D3D3}Baseline
& \cellcolor[HTML]{D3D3D3}39.10 & \cellcolor[HTML]{D3D3D3}1.00 & \cellcolor[HTML]{D3D3D3}3.45 \textcolor[HTML]{D35400}{(1.0$\times$)} \\
& & \cellcolor[HTML]{D3D3D3}Baseline+Parallel
& \cellcolor[HTML]{D3D3D3}38.90 & \cellcolor[HTML]{D3D3D3}1.39 & \cellcolor[HTML]{D3D3D3}20.40 \textcolor[HTML]{D35400}{(5.9$\times$)} \\
& & DAWN
& 37.60 & 1.51 & 27.80 \textcolor[HTML]{D35400}{(8.1$\times$)} \\
& & KLASS
& 30.00 & 1.28 & 13.25 \textcolor[HTML]{D35400}{(3.8$\times$)} \\
& & \cellcolor[HTML]{D5E8D4}\textbf{Polestar (Ours)}
& \cellcolor[HTML]{D5E8D4}\textbf{39.34} & \cellcolor[HTML]{D5E8D4}\textbf{2.10} & \cellcolor[HTML]{D5E8D4}\textbf{48.66} \textcolor[HTML]{D35400}{\textbf{(14.1$\times$)}} \\

\Xhline{2\arrayrulewidth}
\end{tabular}}
\label{tab:llada15_results}
\vspace{-4mm}
\end{wraptable}

\textbf{Comparison with KV Caching Baselines. }In \autoref{tab:final_results}, we compare Polestar against several SoTA dLLM KV caching baselines across different models, benchmarks and generation lengths. Evidently, \textbf{Polestar} achieves the \textbf{best overall accuracy--throughput trade-off, improving accuracy by up to $\mathbf{10.71\%}$} over SoTA baselines, while delivering \textbf{up to $20.5\times$ speedup} over baseline model. Polestar's token-representation-drift-based scheme keeps the KV cache refreshed through selective updates while reliably identifying commit-ready tokens, which translates into high TPF of up to $3.67$. Fast-dLLM \cite{wu2025fast}, which assumes that cached KV states remain static while a block is active, does not account for drift induced by newly decoded tokens, causing consistently lower accuracy and reduced TPF due to degraded prediction confidence. Similarly, Elastic-Cache \cite{nguyen2025attention} and Dynamic-dLLM \cite{wudynamic} are prone to missing broader contextual shifts or triggering unnecessary refreshes. This leads to a weaker accuracy--throughput trade-off than Polestar. 

Importantly, in several settings (e.g., GSM8K-512 and HumanEval-512), \textbf{Polestar improves accuracy over the baseline by up to} $3.63\%$ while also improving throughput. We attribute this to \textbf{Polestar-Commit}, which uses token representation drift to identify commit-ready tokens more reliably than the baseline, where a fixed number of tokens are forced to be committed.

\begin{wraptable}{R}{0.35\textwidth}
\vspace{-5mm}
\begin{minipage}{\linewidth}
\centering
\caption{Accuracy-performance comparison on ParallelBench-\textit{easy} \cite{kang2025parallelbench} using LLaDA-8B-Instruct.}
\vspace{-2mm}
\renewcommand*{\arraystretch}{1.0}
\setlength\tabcolsep{3pt}
\resizebox{\linewidth}{!}{%
\begin{tabular}{l|c|c|c}
\Xhline{2\arrayrulewidth}
Method & Acc(\%)$\uparrow$ & TPF$\uparrow$ & TPS$\uparrow$ \\
\Xhline{2\arrayrulewidth}

\cellcolor[HTML]{D3D3D3}Baseline (LLaDA)
& \cellcolor[HTML]{D3D3D3}77.85
& \cellcolor[HTML]{D3D3D3}1.00
& \cellcolor[HTML]{D3D3D3}10.56 \textcolor[HTML]{D35400}{(1.0$\times$)} \\

\cellcolor[HTML]{D3D3D3}Baseline+Parallel
& \cellcolor[HTML]{D3D3D3}76.08
& \cellcolor[HTML]{D3D3D3}2.31
& \cellcolor[HTML]{D3D3D3}42.23 \textcolor[HTML]{D35400}{(4.0$\times$)} \\

Fast-dLLM
& 68.09 & 1.88 & 49.84 \textcolor[HTML]{D35400}{(4.7$\times$)} \\

Elastic-Cache
& 69.84 & 1.96 & 54.12 \textcolor[HTML]{D35400}{(5.1$\times$)} \\

d$^2$Cache
& 70.42 & 2.05 & 53.42 \textcolor[HTML]{D35400}{(5.1$\times$)} \\

EntropyCache
& 72.55 & 2.18 & 64.88 \textcolor[HTML]{D35400}{(6.1$\times$)} \\

Dynamic-dLLM
& 69.21 & 1.92 & 51.45 \textcolor[HTML]{D35400}{(4.9$\times$)} \\

\cellcolor[HTML]{D5E8D4}\textbf{Polestar (Ours)}
& \cellcolor[HTML]{D5E8D4}\textbf{74.18}
& \cellcolor[HTML]{D5E8D4}\textbf{2.43}
& \cellcolor[HTML]{D5E8D4}\textbf{82.41} \textcolor[HTML]{D35400}{\textbf{(7.8$\times$)}} \\

\Xhline{2\arrayrulewidth}
\end{tabular}}
\label{tab:parallelbench}
\end{minipage}
\vspace{-4mm}
\end{wraptable}
\noindent
\textbf{Comparison with Parallel-Decoding Baselines.}
\autoref{tab:llada15_results} compares Polestar against state-of-the-art parallel-decoding baselines, DAWN~\cite{luo2026dawn} and KLASS~\cite{kim2025klass}, on LLaDA-1.5. Polestar improves accuracy by up to $9.34\%$, achieves TPF up to $3.40$, and delivers up to $3.69\times$ higher speedup over these methods. DAWN \cite{luo2026dawn} achieves substantially lower TPS due to the overhead of constructing pairwise token dependencies during decoding and the absence of KV caching. KLASS \cite{kim2025klass} suffers larger accuracy degradation because its aggressive decoding policy can commit tokens across arbitrary positions. In contrast, Polestar uses a drift-guided commitment policy localized to the current and suffix block, enabling more reliable commitment.   

\textbf{Accuracy and Performance Comparison on Hard-to-Parallelize Tasks. }\autoref{tab:parallelbench} compares Polestar against several baselines on ParallelBench~\cite{kang2025parallelbench}, a benchmark designed to stress-test the accuracy--parallelism trade-off in dLLMs. Since ParallelBench uses shorter generations, we set the generation length to $32$ and block size to $B=8$. \autoref{tab:parallelbench} reports results on the \textit{easy} subset; results on the \textit{hard} subset are provided in \Cref{sec:appendix_parallel_bench}. Polestar remains robust on this benchmark, achieving the highest TPF of $2.43$ and $82.41$ TPS ($7.8\times$ over baseline), while maintaining $74.18\%$ accuracy.



\subsection{Discussions and Ablations}
\vspace{-2mm}
\label{sec:ablations}

\noindent



\newcommand{\poschg}[1]{\textcolor[HTML]{2E7D32}{\scriptsize\,(+#1\%)}}
\newcommand{\posplus}[1]{\textcolor[HTML]{2E7D32}{\scriptsize\,(+#1)}}
\newcommand{\negchg}[1]{\textcolor[HTML]{C00000}{\scriptsize\,(-#1\%)}}
\newcommand{\speedup}[1]{\textcolor[HTML]{D35400}{\scriptsize\,(#1$\times$)}}

\begin{wraptable}{r}{0.50\textwidth}
\vspace{-5.5mm}
\centering
\caption{
Component ablation on LLaDA-8B-Instruct, GSM8K, generation length 256.
}
\label{tab:component_ablation}
\footnotesize
\renewcommand*{\arraystretch}{1.08}
\setlength{\tabcolsep}{3.2pt}
\resizebox{\linewidth}{!}{%
\begin{tabular}{l|c|c|c}
\Xhline{2\arrayrulewidth}
Variant & Acc.(\%)$\uparrow$ & TPF$\uparrow$ & TPS$\uparrow$ \\
\Xhline{2\arrayrulewidth}

\cellcolor[HTML]{D3D3D3}Baseline
& \cellcolor[HTML]{D3D3D3}79.30
& \cellcolor[HTML]{D3D3D3}1.00
& \cellcolor[HTML]{D3D3D3}6.58 \speedup{1.0} \\

\cellcolor[HTML]{D3D3D3}Baseline+Parallel
& \cellcolor[HTML]{D3D3D3}78.57
& \cellcolor[HTML]{D3D3D3}2.95
& \cellcolor[HTML]{D3D3D3}17.65 \speedup{2.7} \\

Fast-dLLM
& 77.88 \poschg{0.00}
& 2.72
& 47.88 \speedup{7.3} \\

\Xhline{1\arrayrulewidth}

\cellcolor[HTML]{D5E8D4}\textbf{Polestar-Cache}
& \cellcolor[HTML]{D5E8D4}78.89\poschg{1.01}
& \cellcolor[HTML]{D5E8D4}2.87\posplus{0.15}
& \cellcolor[HTML]{D5E8D4}60.54 \speedup{9.2} \\
\Xhline{1\arrayrulewidth}
\cellcolor[HTML]{D5E8D4}\textbf{Polestar-Commit}
& \cellcolor[HTML]{D5E8D4}
& \cellcolor[HTML]{D5E8D4}
& \cellcolor[HTML]{D5E8D4} \\

\cellcolor[HTML]{D5E8D4}\textbf{+ Current-Block Commit}
& \cellcolor[HTML]{D5E8D4}78.57\poschg{0.69}
& \cellcolor[HTML]{D5E8D4}3.48\posplus{0.61}
& \cellcolor[HTML]{D5E8D4}82.14 \speedup{12.5} \\

\cellcolor[HTML]{D5E8D4}\textbf{+ Suffix-Block Commit}
& \cellcolor[HTML]{D5E8D4}\textbf{78.33}\poschg{0.45}
& \cellcolor[HTML]{D5E8D4}\textbf{3.67}\posplus{0.19}
& \cellcolor[HTML]{D5E8D4}\textbf{87.57} \textcolor[HTML]{D35400}{\scriptsize\textbf{(13.3$\times$)}} \\

\Xhline{2\arrayrulewidth}
\end{tabular}}
\vspace{-4mm}
\end{wraptable}
\textbf{Impact of Polestar Components.}
\autoref{tab:component_ablation} evaluates the contribution of each Polestar component. Since Polestar-Cache extends Fast-dLLM~\cite{wu2025fast}, we use it as the starting point. Polestar-Cache's drift-aware layer-wise KV refresh policy improves accuracy by $1.01\%$, increases TPF by $0.15$ (reinforces prediction confidence), and improves throughput by $1.26\times$. The throughput gain comes from fewer full-sequence refreshes and system optimizations (see \Cref{sec:appendix_performance_breakdown}). Furthermore, incorporating current-block commitment strategy of Polestar-Commit raises TPF from $2.87$ to $3.48$ with a negligible drop in accuracy and $1.72\times$ throughput improvement over Fast-dLLM. Finally, suffix-block commitment yields Polestar's complete result of $78.33\%$ accuracy, $3.67$ TPF, and $1.83\times$ higher TPS.

\begin{figure}[t]
  \centering
  \makebox[\textwidth][c]{
    \includegraphics[width=0.9\linewidth, keepaspectratio]{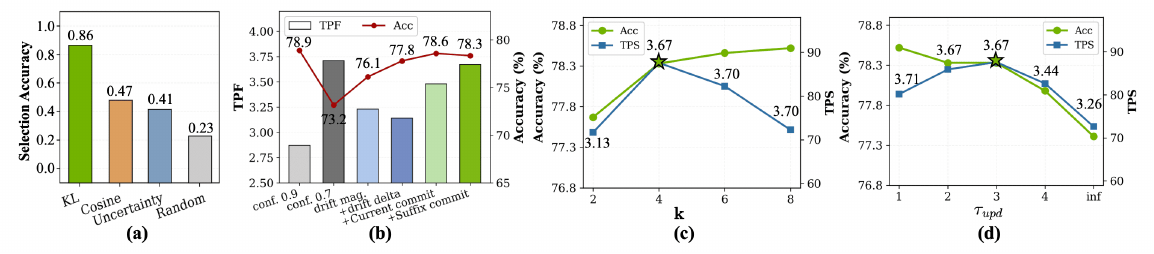}
  }
  \vspace{-1mm}
  \caption{
  Polestar ablations on LLaDA-8B-Instruct, GSM8K, generation length 256 and block size 32
  for (a) drift-policy, (b) commitment-rule, (c) top-$k$ cluster selection and (d) update schedule. For (c,d) star marks the default Polestar setting, and text annotations indicate TPF.
}
  \vspace{-6mm}
  \label{fig:ablations}
\end{figure}

\textbf{Drift Measurement Policy.}
We evaluate Polestar's KL-based drift policy for identifying stale KV-cache positions against most-attended-token cosine similarity~\cite{nguyen2025attention}, uncertainty~\cite{jiang2025d}, and random selection, using the same cached trajectory and refresh budget within Polestar's local window. Each policy selects the top-$25\%$ token positions per layer, and we measure selection accuracy against the true top-$25\%$ most-stale KV-cache positions. For a fair comparison with Elastic-Cache, we compute the cosine-similarity change of each candidate position's most-attended token and use it to rank positions for refresh. As shown in~\cref{fig:ablations}(a), Polestar's KL-based drift achieves substantially higher selection accuracy ($86\%$), indicating that it better identifies KV-cache positions requiring refresh.

\textbf{Commitment Rule.}
For this experiment, we study the impact of different commitment rules, applied on top of Polestar-Cache. As shown in~\cref{fig:ablations}(b), static confidence thresholding with $\tau_s=0.9$ is accurate but conservative, committing only $2.87$ tokens per forward. Lowering the threshold to $\tau_s=0.7$ increases TPF to $3.71$ but causes a large accuracy drop, showing that confidence-only scaling is suboptimal. We therefore fix $\tau_s=0.9$ and next evaluate commitment based on raw drift magnitude and drift delta, where drift delta is measured relative to the mean drift over the past $h$ steps. While raw drift improves over the lowered confidence threshold ($\tau_s=0.7$), drift delta performs better by capturing the desired sharp drift events rather than absolute drift magnitude alone. Polestar-Commit's current block commit policy improves accuracy to $78.6\%$ at $3.48$ TPF by employing a confidence-conditioned dynamic threshold. Finally, suffix commitment increases TPF to $3.67$ with similar accuracy, showing that Polestar-Commit improves decoding parallelism without the collapse caused by naive confidence-threshold lowering.

\textbf{\# of top-$\textbf{k}$ Clusters.}
We vary the number of refreshed clusters while fixing the total number of centroids to $K=8$. As shown in~\cref{fig:ablations}(c), refreshing only the top-2 clusters overlooks token positions that need updates, reaching only $77.67\%$ accuracy and $3.13$ TPF. Increasing the refresh budget improves accuracy but introduces additional computation overhead, reducing throughput for top-6 and top-8. The top-$k=4$ budget achieves $78.33\%$ accuracy, $3.67$ TPF, and TPS of $87.57$, yielding the best tradeoff between accuracy and performance.

\textbf{Update Schedule $\tau_\mathrm{upd}$.}
We vary the selective refresh threshold $\tau_{\mathrm{upd}}$ in~\cref{fig:ablations}(d). Lower thresholds trigger frequent refreshes, preserving accuracy but wasting computation. In contrast, disabling refresh with $\tau_{\mathrm{upd}}=\texttt{inf}$ allows stale states to accumulate, reducing both accuracy and TPF. The $\tau_{\mathrm{upd}}=3$ maintains $78.33\%$ accuracy and $3.67$ TPF while achieving the highest TPS.

\begin{wrapfigure}{R}{0.40\textwidth}
\vspace{-6mm}
  \centering
  \includegraphics[width=0.40\textwidth, keepaspectratio]{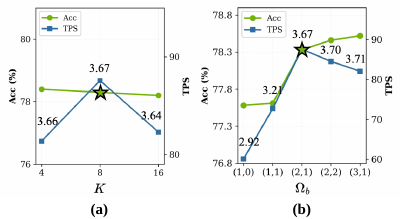}
  \vspace{-5mm}
  \caption{(a) Number of centroids $K$.
  (b) (Prefix,Suffix) local-window size $\Omega_b$.}
  \vspace{-7mm}
  \label{fig:sweep_b}
\end{wrapfigure}

\textbf{\# of centroids $K$. }As shown in~\autoref{fig:sweep_b}(a), $K=8$ provides the best balance between centroid representation granularity and the overhead of computing proxy attention and drift over additional centroids.

\textbf{Local window size $\Omega_b$.}
In \autoref{fig:sweep_b}(b), we ablate different prefix and suffix window sizes of the local window. We observe that expanding the local window, particularly on the prefix side, improves accuracy, TPF, and TPS, since prefix tokens are more impacted by contextual integration than distant masked suffix tokens \cite{wu2025fast}. We therefore adopt $(|\Omega^p_b|,|\Omega^s_b|)=(2B,B)$.


\section{Conclusions}
\vspace{-2mm}
We introduce {Polestar}, a drift-aware inference framework that jointly addresses KV-cache staleness to enable efficient reuse and unreliable parallel token commitment in dLLMs. Polestar leverages token representation drift to selectively refresh stale KV-cache positions and to commit stable tokens earlier in both current and suffix blocks, while system optimizations reduce the overhead of Polestar-specific operations. Across mathematics, coding, and hard-to-parallelize benchmarks, Polestar establishes a new state of the art on the accuracy--throughput Pareto frontier, achieving higher accuracy, throughput, and decoding parallelism than existing dLLM inference baselines.

\bibliography{refs}

@article{nie2025large,
  title={Large language diffusion models},
  author={Nie, Shen and Zhu, Fengqi and You, Zebin and Zhang, Xiaolu and Ou, Jingyang and Hu, Jun and Zhou, Jun and Lin, Yankai and Wen, Ji-Rong and Li, Chongxuan},
  journal={arXiv preprint arXiv:2502.09992},
  year={2025}
}

@article{grattafiori2024llama,
  title={The llama 3 herd of models},
  author={Grattafiori, Aaron and Dubey, Abhimanyu and Jauhri, Abhinav and Pandey, Abhinav and Kadian, Abhishek and Al-Dahle, Ahmad and Letman, Aiesha and Mathur, Akhil and Schelten, Alan and Vaughan, Alex and others},
  journal={arXiv preprint arXiv:2407.21783},
  year={2024}
}

@article{shen2025improving,
  title={Improving the Throughput of Diffusion-based Large Language Models via a Training-Free Confidence-Aware Calibration},
  author={Shen, Jucheng and Sarkar, Gaurav and Ro, Yeonju and Sridhar, Sharath Nittur and Wang, Zhangyang and Akella, Aditya and Kundu, Souvik},
  journal={ACL},
  year={2026}
}

@article{wu2025fast,
  title={Fast-dllm: Training-free acceleration of diffusion llm by enabling kv cache and parallel decoding},
  author={Wu, Chengyue and Zhang, Hao and Xue, Shuchen and Liu, Zhijian and Diao, Shizhe and Zhu, Ligeng and Luo, Ping and Han, Song and Xie, Enze},
  journal={arXiv preprint arXiv:2505.22618},
  year={2025}
}

@article{kang2025parallelbench,
  title={Parallelbench: Understanding the trade-offs of parallel decoding in diffusion llms},
  author={Kang, Wonjun and Galim, Kevin and Oh, Seunghyuk and Lee, Minjae and Zeng, Yuchen and Zhang, Shuibai and Hooper, Coleman and Hu, Yuezhou and Koo, Hyung Il and Cho, Nam Ik and others},
  journal={arXiv preprint arXiv:2510.04767},
  year={2025}
}

@article{ramachandran2025thinkv,
  title={ThinKV: Thought-Adaptive KV Cache Compression for Efficient Reasoning Models},
  author={Ramachandran, Akshat and Neseem, Marina and Sakr, Charbel and Venkatesan, Rangharajan and Khailany, Brucek and Krishna, Tushar},
  journal={arXiv preprint arXiv:2510.01290},
  year={2025}
}

@article{tian2025skipkv,
  title={Skipkv: Selective skipping of kv generation and storage for efficient inference with large reasoning models},
  author={Tian, Jiayi and Azizi, Seyedarmin and Zhao, Yequan and Potraghloo, Erfan Baghaei and McPherson, Sean and Sridhar, Sharath Nittur and Wang, Zhengyang and Zhang, Zheng and Pedram, Massoud and Kundu, Souvik},
  journal={MLSys},
  year={2026}
}

@inproceedings{ramachandran2025microscopiq,
  title={Microscopiq: Accelerating foundational models through outlier-aware microscaling quantization},
  author={Ramachandran, Akshat and Kundu, Souvik and Krishna, Tushar},
  booktitle={Proceedings of the 52nd Annual International Symposium on Computer Architecture},
  pages={1193--1209},
  year={2025}
}

@article{ye2025dream,
  title={Dream 7b: Diffusion large language models},
  author={Ye, Jiacheng and Xie, Zhihui and Zheng, Lin and Gao, Jiahui and Wu, Zirui and Jiang, Xin and Li, Zhenguo and Kong, Lingpeng},
  journal={arXiv preprint arXiv:2508.15487},
  year={2025}
}

@article{arriola2025block,
  title={Block diffusion: Interpolating between autoregressive and diffusion language models},
  author={Arriola, Marianne and Gokaslan, Aaron and Chiu, Justin T and Yang, Zhihan and Qi, Zhixuan and Han, Jiaqi and Sahoo, Subham Sekhar and Kuleshov, Volodymyr},
  journal={arXiv preprint arXiv:2503.09573},
  year={2025}
}

@article{lu2025adablock,
  title={Adablock-dllm: Semantic-aware diffusion llm inference via adaptive block size},
  author={Lu, Guanxi and Chen, Hao Mark and Karashima, Yuto and Wang, Zhican and Fujiki, Daichi and Fan, Hongxiang},
  journal={arXiv preprint arXiv:2509.26432},
  year={2025}
}

@article{nguyen2025attention,
  title={Attention is all you need for kv cache in diffusion llms},
  author={Nguyen-Tri, Quan and Ranjan, Mukul and Shen, Zhiqiang},
  journal={arXiv preprint arXiv:2510.14973},
  year={2025}
}

@article{ma2025dkv,
  title={dkv-cache: The cache for diffusion language models},
  author={Ma, Xinyin and Yu, Runpeng and Fang, Gongfan and Wang, Xinchao},
  journal={arXiv preprint arXiv:2505.15781},
  year={2025}
}

@article{liu2025dllmcache,
  title={dllm-cache: Accelerating diffusion large language models with adaptive caching},
  author={Liu, Zhiyuan and Yang, Yicun and Zhang, Yaojie and Chen, Junjie and Zou, Chang and Wei, Qingyuan and Wang, Shaobo and Zhang, Linfeng},
  journal={arXiv preprint arXiv:2506.06295},
  year={2025}
}

@article{austin2021structured,
  title={Structured denoising diffusion models in discrete state-spaces},
  author={Austin, Jacob and Johnson, Daniel D and Ho, Jonathan and Tarlow, Daniel and Van Den Berg, Rianne},
  journal={Advances in neural information processing systems},
  volume={34},
  pages={17981--17993},
  year={2021}
}

@article{yuan2025native,
  title={Native sparse attention: Hardware-aligned and natively trainable sparse attention},
  author={Yuan, Jingyang and Gao, Huazuo and Dai, Damai and Luo, Junyu and Zhao, Liang and Zhang, Zhengyan and Xie, Zhenda and Wei, YX and Wang, Lean and Xiao, Zhiping and others},
  journal={arXiv preprint arXiv:2502.11089},
  year={2025}
}

@inproceedings{
    qi2026hierarchy,
    title={Hierarchy Decoding: A Training-free Parallel Decoding Strategy  for Diffusion Large Language Models},
    author={Xiaojing Qi and Lun Du and Xinyuan Zhang and Lanning Wei and Tao Jin and Da Zheng},
    booktitle={The Fourteenth International Conference on Learning Representations},
    year={2026},
    url={https://openreview.net/forum?id=ZsIQUjQtdW},
    note={under review}
}

@misc{li2025prophet,
      title={Diffusion Language Models Know the Answer Before Decoding}, 
      author={Pengxiang Li and Yefan Zhou and Dilxat Muhtar and Lu Yin and Shilin Yan and Li Shen and Yi Liang and Soroush Vosoughi and Shiwei Liu},
      year={2025},
      eprint={2508.19982},
      archivePrefix={arXiv},
      primaryClass={cs.CL},
      url={https://arxiv.org/abs/2508.19982}, 
}

@inproceedings{
    wudynamic,
    title={Dynamic-d{LLM}: Dynamic Cache-Budget and Adaptive Parallel Decoding for Training-Free Acceleration of Diffusion {LLM}},
    author={Tianyi Wu and Xiaoxi Sun and Yanhua Jiao and Yulin Li and Yixin Chen and Yun-Hao Cao and Yi-Qi Hu and Zhuotao Tian},
    booktitle={The Fourteenth International Conference on Learning Representations},
    year={2026},
    url={https://openreview.net/forum?id=SdnkB5pGbq}
}

@article{kim2025klass,
  title={KLASS: KL-Guided Fast Inference in Masked Diffusion Models},
  author={Kim, Seo Hyun and Hong, Sunwoo and Jung, Hojung and Park, Youngrok and Yun, Se-Young},
  journal={arXiv preprint arXiv:2511.05664},
  year={2025}
}

@article{luo2026dawn,
  title={DAWN: Dependency-Aware Fast Inference for Diffusion LLMs},
  author={Luo, Lizhuo and Shi, Zhuoran and Luo, Jiajun and Wang, Zhi and Ren, Shen and Wang, Wenya and Zhang, Tianwei},
  journal={arXiv preprint arXiv:2602.06953},
  year={2026}
}

@article{cheong2026entropycache,
  title={EntropyCache: Decoded Token Entropy Guided KV Caching for Diffusion Language Models},
  author={Cheong, Minsoo and Son, Donghyun and Lim, Woosang and Yoo, Sungjoo},
  journal={arXiv preprint arXiv:2603.18489},
  year={2026}
}

@article{hooper2025multipole,
  title={Multipole Attention for Efficient Long Context Reasoning},
  author={Hooper, Coleman and Zhao, Sebastian and Manolache, Luca and Kim, Sehoon and Mahoney, Michael W and Shao, Yakun Sophia and Keutzer, Kurt and Gholami, Amir},
  journal={arXiv preprint arXiv:2506.13059},
  year={2025}
}

@article{cobbe2021training,
  title={Training verifiers to solve math word problems},
  author={Cobbe, Karl and Kosaraju, Vineet and Bavarian, Mohammad and Chen, Mark and Jun, Heewoo and Kaiser, Lukasz and Plappert, Matthias and Tworek, Jerry and Hilton, Jacob and Nakano, Reiichiro and others},
  journal={arXiv preprint arXiv:2110.14168},
  year={2021}
}

@article{jiang2025d,
  title={d2Cache: Accelerating Diffusion-Based LLMs via Dual Adaptive Caching},
  author={Jiang, Yuchu and Cai, Yue and Luo, Xiangzhong and Fu, Jiale and Wang, Jiarui and Liu, Chonghan and Yang, Xu},
  journal={arXiv preprint arXiv:2509.23094},
  year={2025}
}

@misc{
    pan2026blockspec,
    title={BlockSpec: Blockwise Speculative Decoding for Diffusion {LLM}s},
    author={Tianxiang Pan and Baitao Gong and Mo Guang and Hongwei Yong and Tianpeng Jiang and Yaqian Li and Zheng Cao and Kaiwen Long},
    year={2026},
    url={https://openreview.net/forum?id=hmAviop5rm}
}

@misc{wei2025slowfast,
      title={Accelerating Diffusion Large Language Models with SlowFast Sampling: The Three Golden Principles}, 
      author={Qingyan Wei and Yaojie Zhang and Zhiyuan Liu and Dongrui Liu and Linfeng Zhang},
      year={2025},
      eprint={2506.10848},
      archivePrefix={arXiv},
      primaryClass={cs.CL},
      url={https://arxiv.org/abs/2506.10848}, 
}

@misc{song2025sparsedllm,
      title={Sparse-dLLM: Accelerating Diffusion LLMs with Dynamic Cache Eviction}, 
      author={Yuerong Song and Xiaoran Liu and Ruixiao Li and Zhigeng Liu and Zengfeng Huang and Qipeng Guo and Ziwei He and Xipeng Qiu},
      year={2025},
      eprint={2508.02558},
      archivePrefix={arXiv},
      primaryClass={cs.CL},
      url={https://arxiv.org/abs/2508.02558}, 
}

@misc{chen2025dpad,
      title={DPad: Efficient Diffusion Language Models with Suffix Dropout}, 
      author={Xinhua Chen and Sitao Huang and Cong Guo and Chiyue Wei and Yintao He and Jianyi Zhang and Hai "Helen" Li and Yiran Chen},
      year={2025},
      eprint={2508.14148},
      archivePrefix={arXiv},
      primaryClass={cs.CL},
      url={https://arxiv.org/abs/2508.14148}, 
}

@misc{liu2025tidar,
      title={TiDAR: Think in Diffusion, Talk in Autoregression}, 
      author={Jingyu Liu and Xin Dong and Zhifan Ye and Rishabh Mehta and Yonggan Fu and Vartika Singh and Jan Kautz and Ce Zhang and Pavlo Molchanov},
      year={2025},
      eprint={2511.08923},
      archivePrefix={arXiv},
      primaryClass={cs.CL},
      url={https://arxiv.org/abs/2511.08923}, 
}

@misc{yu2025dimple,
      title={Dimple: Discrete Diffusion Multimodal Large Language Model with Parallel Decoding}, 
      author={Runpeng Yu and Xinyin Ma and Xinchao Wang},
      year={2025},
      eprint={2505.16990},
      archivePrefix={arXiv},
      primaryClass={cs.CV},
      url={https://arxiv.org/abs/2505.16990}, 
}

@misc{chen2025ccd,
      title={Beyond Confidence: Adaptive and Coherent Decoding for Diffusion Language Models}, 
      author={Kecheng Chen and Ziru Liu and Xijia Tao and Hui Liu and Xinyu Fu and Suiyun Zhang and Dandan Tu and Lingpeng Kong and Rui Liu and Haoliang Li},
      year={2025},
      eprint={2512.02044},
      archivePrefix={arXiv},
      primaryClass={cs.CL},
      url={https://arxiv.org/abs/2512.02044}, 
}

@misc{zhu2025llada15,
      title={LLaDA 1.5: Variance-Reduced Preference Optimization for Large Language Diffusion Models}, 
      author={Fengqi Zhu and Rongzhen Wang and Shen Nie and Xiaolu Zhang and Chunwei Wu and Jun Hu and Jun Zhou and Jianfei Chen and Yankai Lin and Ji-Rong Wen and Chongxuan Li},
      year={2025},
      eprint={2505.19223},
      archivePrefix={arXiv},
      primaryClass={cs.LG},
      url={https://arxiv.org/abs/2505.19223}, 
}

@misc{you2025lladav,
      title={LLaDA-V: Large Language Diffusion Models with Visual Instruction Tuning}, 
      author={Zebin You and Shen Nie and Xiaolu Zhang and Jun Hu and Jun Zhou and Zhiwu Lu and Ji-Rong Wen and Chongxuan Li},
      year={2025},
      eprint={2505.16933},
      archivePrefix={arXiv},
      primaryClass={cs.LG},
      url={https://arxiv.org/abs/2505.16933}, 
}

@misc{austin2021mbpp,
      title={Program Synthesis with Large Language Models}, 
      author={Jacob Austin and Augustus Odena and Maxwell Nye and Maarten Bosma and Henryk Michalewski and David Dohan and Ellen Jiang and Carrie Cai and Michael Terry and Quoc Le and Charles Sutton},
      year={2021},
      eprint={2108.07732},
      archivePrefix={arXiv},
      primaryClass={cs.PL},
      url={https://arxiv.org/abs/2108.07732}, 
}

@misc{hendrycks2021math,
      title={Measuring Mathematical Problem Solving With the MATH Dataset}, 
      author={Dan Hendrycks and Collin Burns and Saurav Kadavath and Akul Arora and Steven Basart and Eric Tang and Dawn Song and Jacob Steinhardt},
      year={2021},
      eprint={2103.03874},
      archivePrefix={arXiv},
      primaryClass={cs.LG},
      url={https://arxiv.org/abs/2103.03874}, 
}

@misc{chen2021humaneval,
      title={Evaluating Large Language Models Trained on Code}, 
      author={Mark Chen and Jerry Tworek and Heewoo Jun and Qiming Yuan and Henrique Ponde de Oliveira Pinto and Jared Kaplan and Harri Edwards and Yuri Burda and Nicholas Joseph and Greg Brockman and Alex Ray and Raul Puri and Gretchen Krueger and Michael Petrov and Heidy Khlaaf and Girish Sastry and Pamela Mishkin and Brooke Chan and Scott Gray and Nick Ryder and Mikhail Pavlov and Alethea Power and Lukasz Kaiser and Mohammad Bavarian and Clemens Winter and Philippe Tillet and Felipe Petroski Such and Dave Cummings and Matthias Plappert and Fotios Chantzis and Elizabeth Barnes and Ariel Herbert-Voss and William Hebgen Guss and Alex Nichol and Alex Paino and Nikolas Tezak and Jie Tang and Igor Babuschkin and Suchir Balaji and Shantanu Jain and William Saunders and Christopher Hesse and Andrew N. Carr and Jan Leike and Josh Achiam and Vedant Misra and Evan Morikawa and Alec Radford and Matthew Knight and Miles Brundage and Mira Murati and Katie Mayer and Peter Welinder and Bob McGrew and Dario Amodei and Sam McCandlish and Ilya Sutskever and Wojciech Zaremba},
      year={2021},
      eprint={2107.03374},
      archivePrefix={arXiv},
      primaryClass={cs.LG},
      url={https://arxiv.org/abs/2107.03374}, 
}

@misc{lu2024mathvista,
      title={MathVista: Evaluating Mathematical Reasoning of Foundation Models in Visual Contexts}, 
      author={Pan Lu and Hritik Bansal and Tony Xia and Jiacheng Liu and Chunyuan Li and Hannaneh Hajishirzi and Hao Cheng and Kai-Wei Chang and Michel Galley and Jianfeng Gao},
      year={2024},
      eprint={2310.02255},
      archivePrefix={arXiv},
      primaryClass={cs.CV},
      url={https://arxiv.org/abs/2310.02255}, 
}

@misc{zhang2024mathverse,
      title={MathVerse: Does Your Multi-modal LLM Truly See the Diagrams in Visual Math Problems?}, 
      author={Renrui Zhang and Dongzhi Jiang and Yichi Zhang and Haokun Lin and Ziyu Guo and Pengshuo Qiu and Aojun Zhou and Pan Lu and Kai-Wei Chang and Peng Gao and Hongsheng Li},
      year={2024},
      eprint={2403.14624},
      archivePrefix={arXiv},
      primaryClass={cs.CV},
      url={https://arxiv.org/abs/2403.14624}, 
}

@misc{biderman2024lmeval,
      title={Lessons from the Trenches on Reproducible Evaluation of Language Models}, 
      author={Stella Biderman and Hailey Schoelkopf and Lintang Sutawika and Leo Gao and Jonathan Tow and Baber Abbasi and Alham Fikri Aji and Pawan Sasanka Ammanamanchi and Sidney Black and Jordan Clive and Anthony DiPofi and Julen Etxaniz and Benjamin Fattori and Jessica Zosa Forde and Charles Foster and Jeffrey Hsu and Mimansa Jaiswal and Wilson Y. Lee and Haonan Li and Charles Lovering and Niklas Muennighoff and Ellie Pavlick and Jason Phang and Aviya Skowron and Samson Tan and Xiangru Tang and Kevin A. Wang and Genta Indra Winata and François Yvon and Andy Zou},
      year={2024},
      eprint={2405.14782},
      archivePrefix={arXiv},
      primaryClass={cs.CL},
      url={https://arxiv.org/abs/2405.14782}, 
}

@misc{hu2025flashdlm,
      title={FlashDLM: Accelerating Diffusion Language Model Inference via Efficient KV Caching and Guided Diffusion}, 
      author={Zhanqiu Hu and Jian Meng and Yash Akhauri and Mohamed S. Abdelfattah and Jae-sun Seo and Zhiru Zhang and Udit Gupta},
      year={2025},
      eprint={2505.21467},
      archivePrefix={arXiv},
      primaryClass={cs.CL},
      url={https://arxiv.org/abs/2505.21467}, 
}

@misc{israel2025apd,
      title={Accelerating Diffusion LLMs via Adaptive Parallel Decoding}, 
      author={Daniel Israel and Guy Van den Broeck and Aditya Grover},
      year={2025},
      eprint={2506.00413},
      archivePrefix={arXiv},
      primaryClass={cs.CL},
      url={https://arxiv.org/abs/2506.00413}, 
}

@inproceedings{
bao2026learn2pd,
title={Learning to Parallel: Accelerating Diffusion Large Language Models via Learnable Parallel Decoding},
author={Wenrui Bao and Zhiben Chen and Dan Xu and Yuzhang Shang},
booktitle={The Fourteenth International Conference on Learning Representations},
year={2026},
url={https://openreview.net/forum?id=bFJ8Sdr224}
}
\bibliographystyle{icml2026}


\newpage
\appendix
\startcontents[appendix]

\printcontents[appendix]{l}{1}{%
  \section*{\textbf{\huge{Appendix}}}
  \label{sec:appendix}
  \setcounter{tocdepth}{2} 
}
\newpage
\section{Extended Related Works}
\label{appendix_related_work}
Beyond confidence-threshold decoding, CCD~\cite{chen2025ccd} leverages predictive consistency across previous diffusion steps to rectify sampling trajectories and adapt the unmasking budget, while Prophet~\cite{li2025prophet} predicts future denoising trajectories to decide when the model can confidently commit tokens earlier. Speculative and adaptive decoding methods provide another route: BlockSpec~\cite{pan2026blockspec} explores multiple future blockwise trajectories in parallel, APD~\cite{israel2025apd} uses a small auxiliary autoregressive model to adaptively choose how many tokens to decode in parallel, and Learn2PD~\cite{bao2026learn2pd} trains a lightweight filter to predict whether each token prediction should remain unmasked. Other methods modify the decoding structure or model itself: AdaBlock-dLLM~\cite{lu2025adablock} adjusts block sizes using token uncertainty, Hierarchy-dLLM~\cite{qi2026hierarchy} partitions blocks into smaller sub-blocks for more controlled parallelism, SlowFast~\cite{wei2025slowfast} alternates between exploratory and accelerated decoding stages, TiDAR~\cite{liu2025tidar} combines diffusion-style drafting with autoregressive output sampling, and Dimple~\cite{yu2025dimple} introduces a diffusion-trained multimodal LLM with confident decoding. Orthogonal to these parallel-decoding methods, several works reduce per-step cost through cache or attention optimization. FlashDLM~\cite{hu2025flashdlm} combines FreeCache, which reuses stable KV projections, with guided diffusion using a lightweight AR model; dKV-Cache~\cite{ma2025dkv} introduces delayed and conditioned KV caching for denoising; EntropyCache~\cite{cheong2026entropycache} uses decoded-token entropy as a constant-cost skip-or-recompute signal; Sparse-dLLM~\cite{song2025sparsedllm} exploits persistent attention saliency to evict low-relevance prefix/suffix cache entries; and DPad~\cite{chen2025dpad} drops distant suffix tokens to reduce redundant attention computation.

\section{Theoretical Justification of Representation Drift}
\label{sec:theory}

We provide a first-order analysis showing that token representation drift arises from bidirectional attention, even when all hidden states and KV states are recomputed exactly at every denoising step.
This analysis is not an end-to-end performance guarantee for Polestar.
Instead, it formalizes the claim in \Cref{sec:motivation} that drift reflects contextual integration during dLLM inference, rather than merely approximation error introduced by KV-cache reuse.

\paragraph{Setup.}
Consider an attention head at a fixed Transformer layer.
Let $x_j \in \mathbb{R}^{d_{\mathrm{model}}}$ denote the layer input at token position $j$, and let
\[
q_i = W_Q x_i,\qquad
k_j = W_K x_j,\qquad
v_j = W_V x_j
\]
be the query, key, and value vectors.
The attention logit, attention distribution, and attention output for position $i$ are
\[
\ell_{ij} = \frac{q_i^\top k_j}{\sqrt{d_h}},
\qquad
a_{ij}
=
\frac{\exp(\ell_{ij})}{\sum_m \exp(\ell_{im})},
\qquad
o_i = \sum_j a_{ij} v_j ,
\]
where $d_h$ is the head dimension.
We analyze the local effect of unmasking one or more positions on the attention output of another position.
LayerNorm, residual connections, FFN blocks, and multi-layer propagation are omitted from the derivation; they can further propagate the same perturbation, but are not needed to show that bidirectional attention already creates nonzero representation drift under exact recomputation.

\paragraph{Proposition 1. Bidirectional attention induces token representation drift.}
Suppose that between two consecutive denoising steps, position $s$ is unmasked, changing its layer input from $x_s$ to $x_s+\eta$, while the other layer inputs are fixed to first order.
For any position $i\neq s$ that can attend to $s$, the attention output changes as
\[
o_i(x_s+\eta)-o_i(x_s)
=
a_{is} W_V \eta
+
a_{is}
\frac{q_i^\top W_K \eta}{\sqrt{d_h}}
\left(v_s-o_i\right)
+
O(\|\eta\|^2).
\]
Therefore, unmasking position $s$ changes the representation of position $i$ even under exact full-sequence recomputation.

\paragraph{Proof.}
Since $i\neq s$, the query $q_i$ is fixed to first order.
Only the key and value at position $s$ change:
\[
\delta k_s = W_K\eta,\qquad
\delta v_s = W_V\eta.
\]
Thus the only first-order logit perturbation is
\[
\delta \ell_{is}
=
\frac{q_i^\top W_K\eta}{\sqrt{d_h}}.
\]
The softmax derivative gives
\[
\delta a_{ij}
=
a_{ij}
\left(\mathbf{1}\{j=s\}-a_{is}\right)
\delta \ell_{is}.
\]
Therefore,
\[
\begin{aligned}
\delta o_i
&=
\sum_j \delta a_{ij} v_j
+
a_{is}\delta v_s \\
&=
a_{is}\delta \ell_{is}
\left(
v_s-\sum_j a_{ij}v_j
\right)
+
a_{is}W_V\eta \\
&=
a_{is}
\frac{q_i^\top W_K\eta}{\sqrt{d_h}}
\left(v_s-o_i\right)
+
a_{is}W_V\eta .
\end{aligned}
\]
This proves the claimed first-order expression.
In bidirectional attention, a visible position $s$ lies in the attention support of other positions, so $a_{is}>0$ under softmax attention.
Hence the effect is generically nonzero.
In contrast, in causal autoregressive attention, if $s$ is not visible to $i$, then $a_{is}=0$ structurally and this first-order effect vanishes.
\qed

\paragraph{Corollary 1. Hidden-state drift leads to stale KV states under cache reuse.}
Let $\mathbf{H}_i^{(t,\ell)}$ and $\mathbf{H}_i^{(t-1,\ell)}$ denote the exact fully recomputed hidden states at position $i$ and layer $\ell$ across two consecutive denoising steps, and define
\[
\Delta \mathbf{H}_i^{(t,\ell)}
=
\mathbf{H}_i^{(t,\ell)}-\mathbf{H}_i^{(t-1,\ell)}.
\]
The corresponding key and value changes are
\[
\Delta K_i^{(t,\ell)}
=
W_K^{(\ell)} \Delta \mathbf{H}_i^{(t,\ell)},
\qquad
\Delta V_i^{(t,\ell)}
=
W_V^{(\ell)} \Delta \mathbf{H}_i^{(t,\ell)}.
\]
Equivalently,
\[
\left(
\|\Delta K_i^{(t,\ell)}\|_2^2
+
\|\Delta V_i^{(t,\ell)}\|_2^2
\right)^{1/2}
=
\left\|
\begin{bmatrix}
W_K^{(\ell)}\\
W_V^{(\ell)}
\end{bmatrix}
\Delta \mathbf{H}_i^{(t,\ell)}
\right\|_2 .
\]
Thus, if
\[
\Delta \mathbf{H}_i^{(t,\ell)}
\notin
\ker(W_K^{(\ell)}) \cap \ker(W_V^{(\ell)}),
\]
where, $\ker(A)$ is the set of all vectors in the domain that map to the zero vector, i.e., all $x$ such that $Ax=0$. Then, at least one of the exact key or value states changes across steps.
Consequently, reusing the previous KV state without refresh creates a stale cache entry whenever the hidden-state drift is not wiped out by both projections.

Moreover,
\[
\left(
\|\Delta K_i^{(t,\ell)}\|_2^2
+
\|\Delta V_i^{(t,\ell)}\|_2^2
\right)^{1/2}
\le
\left\|
\begin{bmatrix}
W_K^{(\ell)}\\
W_V^{(\ell)}
\end{bmatrix}
\right\|_2
\|\Delta \mathbf{H}_i^{(t,\ell)}\|_2 .
\]
If the combined projection has restricted minimum singular value $\sigma_{\min}>0$ on the drift subspace, then
\[
\left(
\|\Delta K_i^{(t,\ell)}\|_2^2
+
\|\Delta V_i^{(t,\ell)}\|_2^2
\right)^{1/2}
\ge
\sigma_{\min}
\|\Delta \mathbf{H}_i^{(t,\ell)}\|_2 .
\]
Therefore, hidden-state drift generally induces KV-state drift, and approximate KV reuse creates staleness by retaining old KV states after the exact recomputed states have changed.

\paragraph{Proposition 2. KL-based drift bounds attention-output change.}
Let $a=A_i^{(t,\ell)}$ and $a'=A_i^{(t-1,\ell)}$ be the attention distributions of the same token at two consecutive denoising steps.
Assume that the value vectors are fixed for this comparison and satisfy $\|v_j\|_2 \le M$ for all $j$, where $M=\max_j \|v_j\|_2$ over the attention support.
Then
\[
\left\|
\sum_j (a_j-a'_j)v_j
\right\|_2
\le
M \|a-a'\|_1
\le
M\sqrt{2\,\mathrm{KL}(a\|a')}.
\]
Thus, when the KL-based drift between two attention distributions is small, the change in the attention output due solely to attention reweighting is also bounded.
Large KL-based drift indicates that the token's attention distribution has shifted over the full support; whether this shift produces large KV mismatch also depends on the value states, which we evaluate empirically in \cref{sec:results}.

\paragraph{Proof.}
The first inequality follows from the triangle inequality:
\[
\left\|
\sum_j (a_j-a'_j)v_j
\right\|_2
\le
\sum_j |a_j-a'_j|\,\|v_j\|_2
\le
M\|a-a'\|_1 .
\]
The second inequality follows from Pinsker's inequality,
\[
\|a-a'\|_1
\le
\sqrt{2\,\mathrm{KL}(a\|a')}.
\]
\qed

\paragraph{Implication for Polestar.}
Proposition~1 shows that token representation drift arises due to 
bidirectional attention, before any KV-cache approximation is introduced.
Corollary~1 shows that such drift leads to stale KV states when old KV states are reused without refresh. Proposition~2 explains why the KL-based drift in \Cref{equation:drift} is tied to changes in the attention distribution over the full support.
These results justify using token representation drift as a principled signal, while the empirical analyses in \Cref{sec:motivation} and \Cref{sec:ablations} evaluate whether KL-based drift is an effective selector for sparse KV refresh and token commitment.

\section{Supplementary Details on Polestar}
\subsection{Polestar Algorithm}
\label{sec:appendix_polestar_algorithm}
\Cref{alg:polestar} outlines the Polestar procedure, highlighting its three key contributions. For readability, we omit some superscripts, indexing etc. when clear from context to better emphasize the execution flow.

\subsection{KV-Cache Staleness Metric}
\label{sec:appendix_cosine_distance}

For token index $i$ at step $t$ and layer $\ell$, we measure KV cache staleness by the average cosine distance between cached and fully recomputed key-value representations:
\[
E_i^{(t,\ell)}
\triangleq
1 - \frac{1}{2}\Big(
\texttt{cos\_sim}({K}_{[i]}^{(t,\ell)}, \mathbf{K}_{[i],\mathrm{ref}}^{(t_b,\ell)})
+
\texttt{cos\_sim}({V}_{[i]}^{(t,\ell)}, \mathbf{V}_{[i],\mathrm{ref}}^{(t_b,\ell)})
\Big).
\]
Cosine similarity is computed independently for each attention head and averaged across heads.

\subsection{Quantization Format}
\label{sec:appendix_quantization}
Polestar stores cached hidden-state residuals using NVFP4. NVFP4 represents each value with 1 sign bit, 2 exponent bits, and 1 mantissa bit, and uses a group-wise FP8 (E4M3) scale factor with group size 16.

\section{Additional Evaluations}
\begin{table*}[t]
\centering
\caption{Default hyperparameters and sweep ranges used for Polestar.}
\renewcommand*{\arraystretch}{1.1}
\setlength{\tabcolsep}{5pt}
\resizebox{0.8\linewidth}{!}{%
\begin{tabular}{l|c|c}
\Xhline{2\arrayrulewidth}
Hyperparameter & \textbf{Default} & Sweep range \\
\Xhline{2\arrayrulewidth}

Block size ${B}$
& \cellcolor[HTML]{D5E8D4}$\textbf{32}$
& $\{16,32,64\}$ \\

Local window $(|\Omega^p_b|,|\Omega^s_b|)$
& \cellcolor[HTML]{D5E8D4}$\textbf{(2B,1B)}$
& $\{(1,0),(1,1),(2,1),(2,2),(3,1)\}$ \\

Total centroids $K$
& \cellcolor[HTML]{D5E8D4}$\textbf{8}$
& $\{4,8,16\}$ \\

Selected clusters $\textbf{k}$
& \cellcolor[HTML]{D5E8D4}$\textbf{4}$
& $\{2,4,6,8\}$ with $K=8$ \\

Refresh trigger $\tau_{\mathrm{upd}}$
& \cellcolor[HTML]{D5E8D4}$\textbf{3}$
& $\{1,2,3,4,\infty\}$ \\

Commit confidence $\tau_s$
& \cellcolor[HTML]{D5E8D4}$\textbf{0.9}$
& $\{0.7,0.9\}$ \\

Drift history size $\textbf{h}$
& \cellcolor[HTML]{D5E8D4}$\textbf{5}$
& $\{0,1,3,5,7\}$ \\

Drift-gate coefficient $\alpha$
& \cellcolor[HTML]{D5E8D4}$\textbf{10}$
& $\{2.5,5,10,20,40\}$ \\

Suffix confidence
& \cellcolor[HTML]{D5E8D4}$\textbf{0.9}$
& $\{0.85,0.90,0.95\}$ \\

Hidden-state quantization
& \cellcolor[HTML]{D5E8D4}\textbf{Residual NVFP4}
& FP16, FP8, INT8, MXFP4, INT2, NVFP4 \\

\Xhline{2\arrayrulewidth}
\end{tabular}}
\label{tab:appendix_hyperparam}
\end{table*}
\subsection{Extended Experimental Details}
\label{sec:appendix_experiments}
\textbf{Implementation.}
Polestar is implemented in PyTorch by extending a Fast-dLLM-style~\cite{wu2025fast} blockwise parallel decoding and dual-cache inference pipeline. Our full implementation adds (i) Polestar-Cache, drift-aware sparse cache realignment via clustered hidden-state, and (ii) Polestar-Commit, including current-block and suffix-block drift-aware commitment. All system optimizations are implemented in Triton.

\textbf{Performance Measurement. }
For all methods, TPF and TPS are measured over the full fixed-length denoising trajectory.
We do not stop decoding, truncate the internal sequence, or shorten the remaining generation budget when an \texttt{[EOS]} token is first produced.
Instead, decoding continues until all response positions in the prescribed generation length are unmasked.
This is important for dLLMs because, under bidirectional attention, positions beyond an early \texttt{[EOS]} remain part of the denoising context and can influence the representations used to unmask the remaining positions.
For throughput accounting, the generated-token count excludes \texttt{[EOS]} tokens and prompt tokens.
TPF is computed as the number of non-\texttt{[EOS]} generated response tokens divided by the number of model forward passes, and TPS uses the same token count divided by the end-to-end wall-clock decoding time.

\textbf{Hyperparameters.}
\autoref{tab:appendix_hyperparam} summarizes the default hyperparameters used in the main experiments and the sweep ranges used in ablations. Unless otherwise specified, all results use the default configuration. Importantly, Polestar does not require per-model or per-dataset fine-tuning.

\begin{algorithm}[htbp]
\caption{Polestar}
\label{alg:polestar}
\small

\KwIn{
Prompt $\mathbf{p}$, mask token ${[\mathtt{MASK}]}$, maximum diffusion steps $\mathbf{T}$, \# of blocks $\mathbf{\mathcal{B}}$, \# of layers $\mathbf{L}$, local prefix window size $|\mathbf{\Omega^p_{b}}|$, local suffix window size $|\mathbf{\Omega^s_{b}}|$, drift history size $\mathbf{h}$, total \# of centroids $K$, \# of selected centroids $\mathbf{k}$, static confidence threshold $\mathbf{\tau_s}$, drift-gate coefficient $\alpha $
}
\KwOut{Decoded sequence $\mathbf{y^{(T)}}$}

\KwInit{\parbox[t]{0.85\linewidth}{
$y^{(0)} \leftarrow [\,p;\,[\mathtt{MASK}],\ldots,[\mathtt{MASK}]\,],\quad t \leftarrow 0$\\[2pt]
{Cached state:}$\mathbf{C}=\{\mathbf{KV},\mathbf{H},\mathbf{Z},\mathbf{P}\} \leftarrow \varnothing$
\bluecommenttext{$\mathbf{H}$: hidden-state, $\mathbf{Z}$: centroid, $\mathbf{P}$: proxy-attn}\\
$\Theta \leftarrow \varnothing$
\bluecommenttext{Drift window (size $h$)}\\
${U}^{(t,\ell)} \leftarrow \varnothing$
\bluecommenttext{Layer update packet}
}}
\While{$t < T$}{
\For{$b=0$ \KwTo $\mathcal{B}-1$}{

Let $\mathcal{J}_b$ be the token indices of active block $b$

Let $\Omega_b = \{j : j \in [b-|{\Omega^p_{b}}|,\, b+|{\Omega^s_{b}}|],\; j \notin \mathcal{J}_b\}$\;

$E_b \leftarrow 
\begin{cases}
\mathcal{I}, & b \% 2 == 0 \\
\Omega_b \cup \mathcal{J}_b, & \text{otherwise}
\end{cases}$ \bluecommenttext{Set block entry context}

Forward on $y^{(t)}_{{E}_b}$ with cached $\mathbf{KV}_{\bar{E}_b}$

$\mathbf{Z}_{\Omega_b},\mathbf{M}_{\Omega_b} \leftarrow \mathtt{SphericalKMeans}(\mathbf{H}_{\Omega_b}, K)$  \bluecommenttext{Obtain centroids and cluster labels}

$\mathbf{H'}_{\Omega_b} \leftarrow \mathtt{ResidualQuantization}(\mathbf{H}_{\Omega_b}, \mathbf{Z}_{\Omega_b})$ \bluecommenttext{$\forall$ layers}

$\mathbf{P} \leftarrow \mathtt{ProxyAttn}(\mathbf{Z}_{\Omega_b})$ \bluecommenttext{$\forall$ layers}

$\mathbf{C} \leftarrow \mathrm{Update}\!\left(\mathbf{KV}_{E_b \setminus \mathcal{J}_b}, \mathbf{H'}_{\Omega_b}, \mathbf{Z}_{\Omega_b}, \mathbf{P}_{\Omega_b}\right)$ \bluecommenttext{$\forall$ layers}

$c_{\mathcal{J}_b} \gets \mathtt{Confidence}\!\left(y_{\mathcal{J}_b}^{(t)} \mid [\mathtt{MASK}]\right)$

Commit $\{ j \in \mathcal{J}_b : c_j > \tau_s,\; y_j^{(t)} == [\mathtt{MASK}] \}$

$t \leftarrow t + 1$

\While{$\exists\, j \in \mathcal{J}_b : y_j^{(t)} == [\mathtt{MASK}]$}
{
    
    \For{$\ell = 0$ \KwTo $L-1$}{

    $\mathbf{H}_{\Omega_b}^{(t-1,\ell)} \leftarrow \mathtt{Dequantize}(\mathbf{H'}_{\Omega_b}, \mathbf{Z}_{\Omega_b})^{(t-1,\ell)}$ 
    
    $(\mathbf{KV}_{\Omega_b}, \mathbf{H}_{\Omega_b}, \mathbf{Z}_{\Omega_b})^{(t,\ell)} \leftarrow \mathtt{Patch}\!\left((\mathbf{KV}_{\Omega_b}, \mathbf{H}_{\Omega_b}, \mathbf{Z}_{\Omega_b})^{(t-1,\ell)}, \exists{U}^{(t,\ell)}\right)$\;

    Forward on $y^{(t)}_{\mathcal{J}_b}$ using the patched states; compute main attention scores $\mathbf{A}_{\mathcal{J}_b}^{(t,\ell)}$
    
    $\mathbf{P}^{(t,\ell)} \leftarrow \mathtt{ProxyAttn}(\mathbf{Z}^{(t,\ell)}_{\Omega_b})$\;
    
    ${D}^{(t,\ell)} \leftarrow \mathtt{KL}\!\left(\mathbf{P}^{(t,\ell)} \,\|\, \mathbf{P}^{(t-1,\ell)}\right)$ \bluecommenttext{Token drift}
    
    $\mathcal{K} \leftarrow \mathtt{TopK}(\mathbf{D}^{(t,\ell)}, k)$ \bluecommenttext{Identify TopK clusters} 

    $S_{\Omega_b}\leftarrow \mathbf{M}_{\Omega_b}[\mathcal{K}]$ \bluecommenttext{Token indices corresponding to $\mathcal{K}$}

    Forward on $y^{(t)}_{S_{\Omega_b}}$ using the patched states and obtain ${U}^{(t,\ell+1)}$ \bluecommenttext{Update packet}

    $\mathbf{H'}_{\Omega_b}^{(t, \ell)} \leftarrow \mathtt{ResidualQuantization}(\mathbf{H}_{\Omega_b}^{(t, \ell)}, \mathbf{Z}_{\Omega_b}^{(t, \ell)})$
    
    $\mathbf{C}^{(\ell)} \leftarrow \mathrm{Update}\!\left(\mathbf{KV}_{\Omega_b}, \mathbf{H'}_{\Omega_b}, \mathbf{Z}_{\Omega_b}, \mathbf{P}\right)^{(t,\ell)}$ \bluecommenttext{Update cached states}
}

$c_{\mathcal{J}_b} \gets \mathtt{Confidence}\!\left(y_{\mathcal{J}_b}^{(t)} \mid [\mathtt{MASK}]\right)$\bluecommenttext{Get confidence of masked tokens in current block}

$\mathbf{\delta}^{(t)} \leftarrow \mathtt{KL}\!\left(\mathbf{A}_{\mathcal{J}_b}^{(t,L-1)} \,\|\, \mathbf{A}_{\mathcal{J}_b}^{(t-1,L-1)}\right)$ \bluecommenttext{Compute drift in main attention scores of layer L-1}

$\Delta^{(t)} \leftarrow \mathbf{\delta}^{(t)} - \mathtt{mean}(\Theta)$ 

$\Theta \leftarrow (\Theta \cup \{\boldsymbol{\delta}^{(t)}\}){[-h:]}$ \bluecommenttext{Keep recent $h$}

$\tau_{d}^{(t)} \leftarrow \alpha(\tau_s - c_{\mathcal{J}_b}^{(t)})^2$\
\bluecommenttext{Dynamic drift threshold}

\If{$\exists\, j \in \mathcal{J}_b : c_j > \tau_s$}{
    Commit $\{ j \in \mathcal{J}_b : c_j > \tau_s,\; y_j^{(t)} == [\mathtt{MASK}] \}$\; 
}

\If{$\exists\, j \in \mathcal{J}_b : \Delta^{(t)}_j > \tau_{d,j}^{(t)}$}{
    Commit $\{ j \in \mathcal{J}_b : \Delta^{(t)}_j > \tau_{d,j}^{(t)},\; y_j^{(t)} == [\mathtt{MASK}] \}$
}

$c_{\mathcal{J}_{b+1}} \gets \mathtt{Confidence}\!\left({U}^{(t,L)}_{\mathcal{J}_{b+1}} \mid [\mathtt{MASK}]\right)$\bluecommenttext{$\mathcal{J}_{b+1}$ is the suffix portion of $\Omega_b$}

Commit $\{ j \in \mathrm{Idx}({U}_{\mathcal{J}_{b+1}}^{(t,L)}) : c_j > \tau_s, y_{\mathcal{J}_{b+1}}^{(t)} == [\mathtt{MASK}] \}$ 

$t \leftarrow t + 1$

}
}}
\end{algorithm}

\subsection{Results on MBPP}
\label{sec:appendix_mbpp_results}
\begin{table*}[t]
\centering
\caption{LLaDA-8B-Instruct and Dream-7B-Instruct on MBPP. Accuracy is pass@1 (\%), values in $\pm$ denote accuracy error.
}
\renewcommand*{\arraystretch}{1.1}
\setlength{\tabcolsep}{4pt}
\resizebox{0.8\textwidth}{!}{%
\begin{tabular}{c l|ccc|ccc}
\Xhline{2\arrayrulewidth}
\multirow{2}{*}{Gen Len} & \multirow{2}{*}{Method}
& \multicolumn{3}{c|}{LLaDA-8B-Instruct}
& \multicolumn{3}{c}{Dream-7B-Instruct} \\
& 
& Acc(\%)$\uparrow$ & TPF$\uparrow$ & TPS$\uparrow$
& Acc(\%)$\uparrow$ & TPF$\uparrow$ & TPS$\uparrow$ \\
\Xhline{2\arrayrulewidth}

\multirow{6}{*}{256}
& \cellcolor[HTML]{D3D3D3}Baseline
& \cellcolor[HTML]{D3D3D3}28.36{\scriptsize$\pm$2.02} & \cellcolor[HTML]{D3D3D3}1.00 & \cellcolor[HTML]{D3D3D3}9.95 \textcolor[HTML]{D35400}{(1.0$\times$)}
& \cellcolor[HTML]{D3D3D3}56.60{\scriptsize$\pm$2.22} & \cellcolor[HTML]{D3D3D3}1.00 & \cellcolor[HTML]{D3D3D3}16.95 \textcolor[HTML]{D35400}{\textbf{(1.0$\times$)}} \\
& \cellcolor[HTML]{D3D3D3}Baseline+Parallel
& \cellcolor[HTML]{D3D3D3}28.20{\scriptsize$\pm$2.01} & \cellcolor[HTML]{D3D3D3}2.76 & \cellcolor[HTML]{D3D3D3}27.74 \textcolor[HTML]{D35400}{(2.8$\times$)}
& \cellcolor[HTML]{D3D3D3}54.87{\scriptsize$\pm$2.23} & \cellcolor[HTML]{D3D3D3}1.58 & \cellcolor[HTML]{D3D3D3}30.55 \textcolor[HTML]{D35400}{(1.8$\times$)} \\
& Fast-dLLM
& 27.51{\scriptsize$\pm$2.00} & 2.22 & 38.63 \textcolor[HTML]{D35400}{(3.9$\times$)}
& 52.80{\scriptsize$\pm$2.23} & 1.38 & 33.04 \textcolor[HTML]{D35400}{(1.9$\times$)} \\
& Dynamic-dLLM
& 27.56{\scriptsize$\pm$1.98} & 2.29 & 40.01 \textcolor[HTML]{D35400}{(4.0$\times$)}
& 52.21{\scriptsize$\pm$2.22} & 1.56 & 36.71 \textcolor[HTML]{D35400}{(2.2$\times$)} \\
& Elastic-Cache
& 30.40{\scriptsize$\pm$2.06} & 2.29 & 41.07 \textcolor[HTML]{D35400}{(4.1$\times$)}
& 45.69{\scriptsize$\pm$2.21} & 1.04 & 20.05 \textcolor[HTML]{D35400}{(1.2$\times$)} \\
& \cellcolor[HTML]{D5E8D4}\textbf{Polestar (Ours)}
& \cellcolor[HTML]{D5E8D4}\textbf{31.63{\scriptsize$\pm$2.08}} & \cellcolor[HTML]{D5E8D4}\textbf{2.81} & \cellcolor[HTML]{D5E8D4}\textbf{55.68} \textcolor[HTML]{D35400}{\textbf{(5.6$\times$)}}
& \cellcolor[HTML]{D5E8D4}\textbf{54.91{\scriptsize$\pm$2.22}} & \cellcolor[HTML]{D5E8D4}\textbf{1.86} & \cellcolor[HTML]{D5E8D4}\textbf{49.88} \textcolor[HTML]{D35400}{\textbf{(2.9$\times$)}} \\
\hhline{--------}

\multirow{6}{*}{512}
& \cellcolor[HTML]{D3D3D3}Baseline
& \cellcolor[HTML]{D3D3D3}14.12{\scriptsize$\pm$1.56} & \cellcolor[HTML]{D3D3D3}1.00 & \cellcolor[HTML]{D3D3D3}6.91 \textcolor[HTML]{D35400}{(1.0$\times$)}
& \cellcolor[HTML]{D3D3D3}53.80{\scriptsize$\pm$2.21} & \cellcolor[HTML]{D3D3D3}1.00 & \cellcolor[HTML]{D3D3D3}10.80 \textcolor[HTML]{D35400}{\textbf{(1.0$\times$)}} \\
& \cellcolor[HTML]{D3D3D3}Baseline+Parallel
& \cellcolor[HTML]{D3D3D3}14.12{\scriptsize$\pm$1.55} & \cellcolor[HTML]{D3D3D3}2.73 & \cellcolor[HTML]{D3D3D3}18.90 \textcolor[HTML]{D35400}{(2.7$\times$)}
& \cellcolor[HTML]{D3D3D3}53.10{\scriptsize$\pm$2.23} & \cellcolor[HTML]{D3D3D3}1.50 & \cellcolor[HTML]{D3D3D3}30.38 \textcolor[HTML]{D35400}{(2.8$\times$)} \\
& Fast-dLLM
& 13.71{\scriptsize$\pm$1.54} & 2.36 & 37.70 \textcolor[HTML]{D35400}{(5.5$\times$)}
& 50.80{\scriptsize$\pm$2.24} & 1.42 & 28.03 \textcolor[HTML]{D35400}{(2.6$\times$)} \\
& Dynamic-dLLM
& 13.71{\scriptsize$\pm$1.54} & 2.95 & 47.11 \textcolor[HTML]{D35400}{(6.8$\times$)}
& 49.70{\scriptsize$\pm$2.24} & 1.47 & 28.44 \textcolor[HTML]{D35400}{(2.6$\times$)} \\
& Elastic-Cache
& 14.16{\scriptsize$\pm$1.55} & 2.38 & 39.13 \textcolor[HTML]{D35400}{(5.7$\times$)}
& 39.62{\scriptsize$\pm$2.19} & 0.94 & 14.21 \textcolor[HTML]{D35400}{(1.3$\times$)} \\
& \cellcolor[HTML]{D5E8D4}\textbf{Polestar (Ours)}
& \cellcolor[HTML]{D5E8D4}\textbf{14.12{\scriptsize$\pm$1.56}} & \cellcolor[HTML]{D5E8D4}\textbf{3.06} & \cellcolor[HTML]{D5E8D4}\textbf{51.38} \textcolor[HTML]{D35400}{\textbf{(7.4$\times$)}}
& \cellcolor[HTML]{D5E8D4}\textbf{51.36{\scriptsize$\pm$2.24}} & \cellcolor[HTML]{D5E8D4}\textbf{1.93} & \cellcolor[HTML]{D5E8D4}\textbf{46.39} \textcolor[HTML]{D35400}{\textbf{(4.3$\times$)}} \\

\Xhline{2\arrayrulewidth}
\end{tabular}}
\label{tab:mbpp_only}
\end{table*}

~\autoref{tab:mbpp_only} reports additional MBPP results for LLaDA-8B-Instruct and Dream-7B-Instruct. Polestar achieves the best overall throughput on both models and both generation lengths while maintaining competitive or stronger pass@1 accuracy. On LLaDA-8B-Instruct, Polestar improves accuracy over the full-recomputation baseline at generation length 256 and matches the baseline at generation length 512, while achieving $5.6\times$ and $7.4\times$ speedup, respectively. On Dream-7B-Instruct, Polestar delivers the highest TPS among all methods and remains substantially more accurate than prior cache-reuse baselines, showing that the proposed drift-based cache realignment is also effective on code-generation workloads.

\subsection{Comparison with d$^2$Cache and EntropyCache}
\label{sec:appendix_entropy_d2_results}
\begin{table*}[t]
\centering
\caption{Extended comparison with d$^2$Cache and EntropyCache across model families.}
\renewcommand*{\arraystretch}{1.1}
\setlength{\tabcolsep}{3pt}
\resizebox{\textwidth}{!}{%
\begin{tabular}{lc l|ccc|ccc|ccc}
\Xhline{2\arrayrulewidth}
\multirow{2}{*}{Benchmark} & \multirow{2}{*}{Gen Len} & \multirow{2}{*}{Method}
& \multicolumn{3}{c|}{LLaDA-8B-Instruct}
& \multicolumn{3}{c|}{LLaDA-1.5}
& \multicolumn{3}{c}{Dream-7B-Instruct} \\
& &
& Acc(\%)$\uparrow$ & TPF$\uparrow$ & TPS$\uparrow$
& Acc(\%)$\uparrow$ & TPF$\uparrow$ & TPS$\uparrow$
& Acc(\%)$\uparrow$ & TPF$\uparrow$ & TPS$\uparrow$ \\
\Xhline{2\arrayrulewidth}

\multirow{10}{*}{\makecell{GSM8K \\ \textit{(5-shot)}}}
& \multirow{5}{*}{256} & \cellcolor[HTML]{D3D3D3}Baseline
& \cellcolor[HTML]{D3D3D3}79.30 & \cellcolor[HTML]{D3D3D3}1.00 & \cellcolor[HTML]{D3D3D3}6.58 \textcolor[HTML]{D35400}{(1.0$\times$)}
& \cellcolor[HTML]{D3D3D3}82.27 & \cellcolor[HTML]{D3D3D3}1.00 & \cellcolor[HTML]{D3D3D3}7.47 \textcolor[HTML]{D35400}{(1.0$\times$)}
& \cellcolor[HTML]{D3D3D3}75.85 & \cellcolor[HTML]{D3D3D3}1.00 & \cellcolor[HTML]{D3D3D3}9.85 \textcolor[HTML]{D35400}{(1.0$\times$)} \\
& & \cellcolor[HTML]{D3D3D3}Baseline+Parallel
& \cellcolor[HTML]{D3D3D3}78.57 & \cellcolor[HTML]{D3D3D3}2.95 & \cellcolor[HTML]{D3D3D3}17.65 \textcolor[HTML]{D35400}{(2.7$\times$)}
& \cellcolor[HTML]{D3D3D3}81.82 & \cellcolor[HTML]{D3D3D3}2.79 & \cellcolor[HTML]{D3D3D3}19.78 \textcolor[HTML]{D35400}{(2.6$\times$)}
& \cellcolor[HTML]{D3D3D3}72.96 & \cellcolor[HTML]{D3D3D3}1.56 & \cellcolor[HTML]{D3D3D3}12.10 \textcolor[HTML]{D35400}{(1.2$\times$)} \\
& & d$^2$Cache
& 77.91 & 2.50 & 46.98 \textcolor[HTML]{D35400}{(7.1$\times$)}
& 79.05 & 2.28 & 41.52 \textcolor[HTML]{D35400}{(5.6$\times$)}
& 70.16 & 1.45 & 39.37 \textcolor[HTML]{D35400}{(4.0$\times$)} \\
& & EntropyCache
& 77.88 & 2.79 & 56.59 \textcolor[HTML]{D35400}{(8.6$\times$)}
& 78.81 & 2.53 & 49.88 \textcolor[HTML]{D35400}{(6.7$\times$)}
& 68.28 & 1.47 & 44.29 \textcolor[HTML]{D35400}{(4.5$\times$)} \\
& & \cellcolor[HTML]{D5E8D4}\textbf{Polestar (Ours)}
& \cellcolor[HTML]{D5E8D4}\textbf{78.33} & \cellcolor[HTML]{D5E8D4}\textbf{3.67} & \cellcolor[HTML]{D5E8D4}\textbf{87.57} \textcolor[HTML]{D35400}{\textbf{(13.3$\times$)}}
& \cellcolor[HTML]{D5E8D4}\textbf{81.06} & \cellcolor[HTML]{D5E8D4}\textbf{3.40} & \cellcolor[HTML]{D5E8D4}\textbf{79.65} \textcolor[HTML]{D35400}{\textbf{(10.7$\times$)}}
& \cellcolor[HTML]{D5E8D4}\textbf{72.40} & \cellcolor[HTML]{D5E8D4}\textbf{2.39} & \cellcolor[HTML]{D5E8D4}\textbf{52.80} \textcolor[HTML]{D35400}{\textbf{(5.4$\times$)}} \\
\hhline{~-----------}

& \multirow{5}{*}{512} & \cellcolor[HTML]{D3D3D3}Baseline
& \cellcolor[HTML]{D3D3D3}77.50 & \cellcolor[HTML]{D3D3D3}1.00 & \cellcolor[HTML]{D3D3D3}3.93 \textcolor[HTML]{D35400}{(1.0$\times$)}
& \cellcolor[HTML]{D3D3D3}82.20 & \cellcolor[HTML]{D3D3D3}1.00 & \cellcolor[HTML]{D3D3D3}3.66 \textcolor[HTML]{D35400}{(1.0$\times$)}
& \cellcolor[HTML]{D3D3D3}76.90 & \cellcolor[HTML]{D3D3D3}1.00 & \cellcolor[HTML]{D3D3D3}6.59 \textcolor[HTML]{D35400}{(1.0$\times$)} \\
& & \cellcolor[HTML]{D3D3D3}Baseline+Parallel
& \cellcolor[HTML]{D3D3D3}77.50 & \cellcolor[HTML]{D3D3D3}2.85 & \cellcolor[HTML]{D3D3D3}18.82 \textcolor[HTML]{D35400}{(4.8$\times$)}
& \cellcolor[HTML]{D3D3D3}81.06 & \cellcolor[HTML]{D3D3D3}2.44 & \cellcolor[HTML]{D3D3D3}15.98 \textcolor[HTML]{D35400}{(4.4$\times$)}
& \cellcolor[HTML]{D3D3D3}69.95 & \cellcolor[HTML]{D3D3D3}1.96 & \cellcolor[HTML]{D3D3D3}15.99 \textcolor[HTML]{D35400}{(2.4$\times$)} \\
& & d$^2$Cache
& 77.04 & 2.39 & 38.50 \textcolor[HTML]{D35400}{(9.8$\times$)}
& 78.31 & 1.94 & 29.07 \textcolor[HTML]{D35400}{(7.9$\times$)}
& 67.07 & 1.44 & 28.69 \textcolor[HTML]{D35400}{(4.4$\times$)} \\
& & EntropyCache
& 76.67 & 2.57 & 42.86 \textcolor[HTML]{D35400}{(10.9$\times$)}
& 78.16 & 2.15 & 32.23 \textcolor[HTML]{D35400}{(8.8$\times$)}
& 67.01 & 1.39 & 29.16 \textcolor[HTML]{D35400}{(4.4$\times$)} \\
& & \cellcolor[HTML]{D5E8D4}\textbf{Polestar (Ours)}
& \cellcolor[HTML]{D5E8D4}\textbf{78.18} & \cellcolor[HTML]{D5E8D4}\textbf{3.59} & \cellcolor[HTML]{D5E8D4}\textbf{80.60} \textcolor[HTML]{D35400}{\textbf{(20.5$\times$)}}
& \cellcolor[HTML]{D5E8D4}\textbf{81.06} & \cellcolor[HTML]{D5E8D4}\textbf{3.70} & \cellcolor[HTML]{D5E8D4}\textbf{61.35} \textcolor[HTML]{D35400}{\textbf{(16.8$\times$)}}
& \cellcolor[HTML]{D5E8D4}\textbf{69.65} & \cellcolor[HTML]{D5E8D4}\textbf{2.13} & \cellcolor[HTML]{D5E8D4}\textbf{34.44} \textcolor[HTML]{D35400}{\textbf{(5.2$\times$)}} \\
\Xhline{1\arrayrulewidth}

\multirow{10}{*}{\makecell{MATH \\ \textit{(4-shot)}}}
& \multirow{5}{*}{256} & \cellcolor[HTML]{D3D3D3}Baseline
& \cellcolor[HTML]{D3D3D3}33.50 & \cellcolor[HTML]{D3D3D3}1.00 & \cellcolor[HTML]{D3D3D3}9.10 \textcolor[HTML]{D35400}{(1.0$\times$)}
& \cellcolor[HTML]{D3D3D3}30.88 & \cellcolor[HTML]{D3D3D3}1.00 & \cellcolor[HTML]{D3D3D3}6.98 \textcolor[HTML]{D35400}{(1.0$\times$)}
& \cellcolor[HTML]{D3D3D3}41.69 & \cellcolor[HTML]{D3D3D3}1.00 & \cellcolor[HTML]{D3D3D3}10.65 \textcolor[HTML]{D35400}{(1.0$\times$)} \\
& & \cellcolor[HTML]{D3D3D3}Baseline+Parallel
& \cellcolor[HTML]{D3D3D3}33.10 & \cellcolor[HTML]{D3D3D3}2.51 & \cellcolor[HTML]{D3D3D3}22.90 \textcolor[HTML]{D35400}{(2.5$\times$)}
& \cellcolor[HTML]{D3D3D3}28.46& \cellcolor[HTML]{D3D3D3}2.25 & \cellcolor[HTML]{D3D3D3}16.79 \textcolor[HTML]{D35400}{(2.4$\times$)}
& \cellcolor[HTML]{D3D3D3}41.30 & \cellcolor[HTML]{D3D3D3}1.68 & \cellcolor[HTML]{D3D3D3}25.87 \textcolor[HTML]{D35400}{(2.4$\times$)} \\
& & d$^2$Cache
& 30.82 & 2.13 & 37.41 \textcolor[HTML]{D35400}{(4.1$\times$)}
& 26.51 & 1.87 & 37.05 \textcolor[HTML]{D35400}{(5.3$\times$)}
& 37.80 & 1.58 & 52.72 \textcolor[HTML]{D35400}{(5.0$\times$)} \\
& & EntropyCache
& 30.26 & 2.43 & 47.73 \textcolor[HTML]{D35400}{(5.2$\times$)}
& 26.25 & 2.14 & 45.61 \textcolor[HTML]{D35400}{(6.5$\times$)}
& 39.69 & 1.67 & 57.51 \textcolor[HTML]{D35400}{(5.4$\times$)} \\
& & \cellcolor[HTML]{D5E8D4}\textbf{Polestar (Ours)}
& \cellcolor[HTML]{D5E8D4}\textbf{32.77} & \cellcolor[HTML]{D5E8D4}\textbf{2.91} & \cellcolor[HTML]{D5E8D4}\textbf{70.05} \textcolor[HTML]{D35400}{\textbf{(7.7$\times$)}}
& \cellcolor[HTML]{D5E8D4}\textbf{27.27} & \cellcolor[HTML]{D5E8D4}\textbf{2.64} & \cellcolor[HTML]{D5E8D4}\textbf{70.84} \textcolor[HTML]{D35400}{\textbf{(10.1$\times$)}}
& \cellcolor[HTML]{D5E8D4}\textbf{40.75} & \cellcolor[HTML]{D5E8D4}\textbf{2.29} & \cellcolor[HTML]{D5E8D4}\textbf{59.62} \textcolor[HTML]{D35400}{\textbf{(5.6$\times$)}} \\
\hhline{~-----------}

& \multirow{5}{*}{512} & \cellcolor[HTML]{D3D3D3}Baseline
& \cellcolor[HTML]{D3D3D3}36.60 & \cellcolor[HTML]{D3D3D3}1.00 & \cellcolor[HTML]{D3D3D3}8.30 \textcolor[HTML]{D35400}{(1.0$\times$)}
& \cellcolor[HTML]{D3D3D3}32.65 & \cellcolor[HTML]{D3D3D3}1.00 & \cellcolor[HTML]{D3D3D3}6.12 \textcolor[HTML]{D35400}{(1.0$\times$)}
& \cellcolor[HTML]{D3D3D3}38.95 & \cellcolor[HTML]{D3D3D3}1.00 & \cellcolor[HTML]{D3D3D3}9.80 \textcolor[HTML]{D35400}{(1.0$\times$)} \\
& & \cellcolor[HTML]{D3D3D3}Baseline+Parallel
& \cellcolor[HTML]{D3D3D3}36.15 & \cellcolor[HTML]{D3D3D3}2.50 & \cellcolor[HTML]{D3D3D3}19.60 \textcolor[HTML]{D35400}{(2.4$\times$)}
& \cellcolor[HTML]{D3D3D3}30.64 & \cellcolor[HTML]{D3D3D3}2.49 & \cellcolor[HTML]{D3D3D3}15.34 \textcolor[HTML]{D35400}{(2.5$\times$)}
& \cellcolor[HTML]{D3D3D3}37.90 & \cellcolor[HTML]{D3D3D3}2.10 & \cellcolor[HTML]{D3D3D3}32.80 \textcolor[HTML]{D35400}{(3.3$\times$)} \\
& & d$^2$Cache
& 34.24 & 2.39 & 49.33 \textcolor[HTML]{D35400}{(5.9$\times$)}
& 31.60 & 1.98 & 36.63 \textcolor[HTML]{D35400}{(6.0$\times$)}
& 34.17 & 2.03 & 57.37 \textcolor[HTML]{D35400}{(5.9$\times$)} \\
& & EntropyCache
& 33.45 & 2.38 & 59.90 \textcolor[HTML]{D35400}{(7.2$\times$)}
& 31.15 & 2.29 & 43.12 \textcolor[HTML]{D35400}{(7.0$\times$)}
& 33.22 & 2.07 & 59.98 \textcolor[HTML]{D35400}{(6.1$\times$)} \\
& & \cellcolor[HTML]{D5E8D4}\textbf{Polestar (Ours)}
& \cellcolor[HTML]{D5E8D4}\textbf{34.85} & \cellcolor[HTML]{D5E8D4}\textbf{3.30} & \cellcolor[HTML]{D5E8D4}\textbf{69.84} \textcolor[HTML]{D35400}{\textbf{(8.4$\times$)}}
& \cellcolor[HTML]{D5E8D4}\textbf{31.77} & \cellcolor[HTML]{D5E8D4}\textbf{2.82} & \cellcolor[HTML]{D5E8D4}\textbf{64.36} \textcolor[HTML]{D35400}{\textbf{(10.5$\times$)}}
& \cellcolor[HTML]{D5E8D4}\textbf{35.71} & \cellcolor[HTML]{D5E8D4}\textbf{2.81} & \cellcolor[HTML]{D5E8D4}\textbf{66.21} \textcolor[HTML]{D35400}{\textbf{(6.8$\times$)}} \\
\Xhline{1\arrayrulewidth}

\multirow{10}{*}{\makecell{HumanEval \\ \textit{(0-shot)}}}
& \multirow{5}{*}{256} & \cellcolor[HTML]{D3D3D3}Baseline
& \cellcolor[HTML]{D3D3D3}43.35 & \cellcolor[HTML]{D3D3D3}1.00 & \cellcolor[HTML]{D3D3D3}17.60 \textcolor[HTML]{D35400}{(1.0$\times$)}
& \cellcolor[HTML]{D3D3D3}53.06 & \cellcolor[HTML]{D3D3D3}1.00 & \cellcolor[HTML]{D3D3D3}5.55 \textcolor[HTML]{D35400}{(1.0$\times$)}
& \cellcolor[HTML]{D3D3D3}58.80 & \cellcolor[HTML]{D3D3D3}1.00 & \cellcolor[HTML]{D3D3D3}23.30 \textcolor[HTML]{D35400}{(1.0$\times$)} \\
& & \cellcolor[HTML]{D3D3D3}Baseline+Parallel
& \cellcolor[HTML]{D3D3D3}43.90 & \cellcolor[HTML]{D3D3D3}3.11 & \cellcolor[HTML]{D3D3D3}56.41 \textcolor[HTML]{D35400}{(3.2$\times$)}
& \cellcolor[HTML]{D3D3D3}52.65 & \cellcolor[HTML]{D3D3D3}2.35 & \cellcolor[HTML]{D3D3D3}13.04 \textcolor[HTML]{D35400}{(2.3$\times$)}
& \cellcolor[HTML]{D3D3D3}58.80 & \cellcolor[HTML]{D3D3D3}2.10 & \cellcolor[HTML]{D3D3D3}46.97 \textcolor[HTML]{D35400}{(2.0$\times$)} \\
& & d$^2$Cache
& 38.36 & 2.98 & 64.05 \textcolor[HTML]{D35400}{(3.6$\times$)}
& 44.85 & 2.18 & 18.41 \textcolor[HTML]{D35400}{(3.3$\times$)}
& 51.83 & 1.45 & 46.84 \textcolor[HTML]{D35400}{(2.0$\times$)} \\
& & EntropyCache
& 40.48 & 2.46 & 56.67 \textcolor[HTML]{D35400}{(3.2$\times$)}
& 47.35 & 2.03 & 16.58 \textcolor[HTML]{D35400}{(3.0$\times$)}
& 54.54 & 1.43 & 49.45 \textcolor[HTML]{D35400}{(2.1$\times$)} \\
& & \cellcolor[HTML]{D5E8D4}\textbf{Polestar (Ours)}
& \cellcolor[HTML]{D5E8D4}\textbf{42.42} & \cellcolor[HTML]{D5E8D4}\textbf{3.38} & \cellcolor[HTML]{D5E8D4}\textbf{87.45} \textcolor[HTML]{D35400}{\textbf{(5.0$\times$)}}
& \cellcolor[HTML]{D5E8D4}\textbf{48.78} & \cellcolor[HTML]{D5E8D4}\textbf{3.58} & \cellcolor[HTML]{D5E8D4}\textbf{70.07} \textcolor[HTML]{D35400}{\textbf{(12.6$\times$)}}
& \cellcolor[HTML]{D5E8D4}\textbf{57.69} & \cellcolor[HTML]{D5E8D4}\textbf{1.79} & \cellcolor[HTML]{D5E8D4}\textbf{55.18} \textcolor[HTML]{D35400}{\textbf{(2.4$\times$)}} \\
\hhline{~-----------}

& \multirow{5}{*}{512} & \cellcolor[HTML]{D3D3D3}Baseline
& \cellcolor[HTML]{D3D3D3}44.10 & \cellcolor[HTML]{D3D3D3}1.00 & \cellcolor[HTML]{D3D3D3}8.57 \textcolor[HTML]{D35400}{(1.0$\times$)}
& \cellcolor[HTML]{D3D3D3}51.84 & \cellcolor[HTML]{D3D3D3}1.00 & \cellcolor[HTML]{D3D3D3}3.53 \textcolor[HTML]{D35400}{(1.0$\times$)}
& \cellcolor[HTML]{D3D3D3}58.80 & \cellcolor[HTML]{D3D3D3}1.00 & \cellcolor[HTML]{D3D3D3}15.48 \textcolor[HTML]{D35400}{(1.0$\times$)} \\
& & \cellcolor[HTML]{D3D3D3}Baseline+Parallel
& \cellcolor[HTML]{D3D3D3}44.10 & \cellcolor[HTML]{D3D3D3}2.71 & \cellcolor[HTML]{D3D3D3}34.53 \textcolor[HTML]{D35400}{(4.0$\times$)}
& \cellcolor[HTML]{D3D3D3}51.21 & \cellcolor[HTML]{D3D3D3}2.23 & \cellcolor[HTML]{D3D3D3}7.01 \textcolor[HTML]{D35400}{(2.0$\times$)}
& \cellcolor[HTML]{D3D3D3}56.95 & \cellcolor[HTML]{D3D3D3}1.93 & \cellcolor[HTML]{D3D3D3}26.77 \textcolor[HTML]{D35400}{(1.7$\times$)} \\
& & d$^2$Cache
& 43.05 & 2.62 & 63.70 \textcolor[HTML]{D35400}{(7.4$\times$)}
& 48.10 & 2.03 & 13.92 \textcolor[HTML]{D35400}{(3.9$\times$)}
& 50.68 & 1.31 & 37.36 \textcolor[HTML]{D35400}{(2.4$\times$)} \\
& & EntropyCache
& 47.39 & 2.23 & 57.80 \textcolor[HTML]{D35400}{(6.7$\times$)}
& 50.15 & 1.66 & 11.75 \textcolor[HTML]{D35400}{(3.3$\times$)}
& 53.35 & 1.39 & 40.27 \textcolor[HTML]{D35400}{(2.6$\times$)} \\
& & \cellcolor[HTML]{D5E8D4}\textbf{Polestar (Ours)}
& \cellcolor[HTML]{D5E8D4}\textbf{47.73} & \cellcolor[HTML]{D5E8D4}\textbf{3.23} & \cellcolor[HTML]{D5E8D4}\textbf{77.67} \textcolor[HTML]{D35400}{\textbf{(9.1$\times$)}}
& \cellcolor[HTML]{D5E8D4}\textbf{50.12} & \cellcolor[HTML]{D5E8D4}\textbf{3.28} & \cellcolor[HTML]{D5E8D4}\textbf{61.95} \textcolor[HTML]{D35400}{\textbf{(17.5$\times$)}}
& \cellcolor[HTML]{D5E8D4}\textbf{54.36} & \cellcolor[HTML]{D5E8D4}\textbf{1.77} & \cellcolor[HTML]{D5E8D4}\textbf{41.68} \textcolor[HTML]{D35400}{\textbf{(2.7$\times$)}} \\
\Xhline{1\arrayrulewidth}

\multirow{10}{*}{\makecell{MBPP \\ \textit{(3-shot)}}}
& \multirow{5}{*}{256} & \cellcolor[HTML]{D3D3D3}Baseline
& \cellcolor[HTML]{D3D3D3}28.36 & \cellcolor[HTML]{D3D3D3}1.00 & \cellcolor[HTML]{D3D3D3}9.95 \textcolor[HTML]{D35400}{(1.0$\times$)}
& \cellcolor[HTML]{D3D3D3}40.02 & \cellcolor[HTML]{D3D3D3}1.00 & \cellcolor[HTML]{D3D3D3}7.87 \textcolor[HTML]{D35400}{(1.0$\times$)}
& \cellcolor[HTML]{D3D3D3}56.60 & \cellcolor[HTML]{D3D3D3}1.00 & \cellcolor[HTML]{D3D3D3}16.95 \textcolor[HTML]{D35400}{(1.0$\times$)} \\
& & \cellcolor[HTML]{D3D3D3}Baseline+Parallel
& \cellcolor[HTML]{D3D3D3}28.20 & \cellcolor[HTML]{D3D3D3}2.76 & \cellcolor[HTML]{D3D3D3}27.74 \textcolor[HTML]{D35400}{(2.8$\times$)}
& \cellcolor[HTML]{D3D3D3}38.44 & \cellcolor[HTML]{D3D3D3}1.57 & \cellcolor[HTML]{D3D3D3}12.36 \textcolor[HTML]{D35400}{(1.6$\times$)}
& \cellcolor[HTML]{D3D3D3}54.87 & \cellcolor[HTML]{D3D3D3}1.58 & \cellcolor[HTML]{D3D3D3}30.55 \textcolor[HTML]{D35400}{(1.8$\times$)} \\
& & d$^2$Cache
& 30.54 & 2.56 & 50.99 \textcolor[HTML]{D35400}{(5.1$\times$)}
& 38.21 & 1.46 & 30.91 \textcolor[HTML]{D35400}{(3.9$\times$)}
& 50.01 & 1.67 & 42.63 \textcolor[HTML]{D35400}{(2.5$\times$)} \\
& & EntropyCache
& 27.37 & 2.80 & 55.01 \textcolor[HTML]{D35400}{(5.5$\times$)}
& 36.42 & 1.44 & 32.55 \textcolor[HTML]{D35400}{(4.1$\times$)}
& 50.03 & 1.68 & 45.78 \textcolor[HTML]{D35400}{(2.7$\times$)} \\
& & \cellcolor[HTML]{D5E8D4}\textbf{Polestar (Ours)}
& \cellcolor[HTML]{D5E8D4}\textbf{31.63} & \cellcolor[HTML]{D5E8D4}\textbf{2.81} & \cellcolor[HTML]{D5E8D4}\textbf{55.68} \textcolor[HTML]{D35400}{\textbf{(5.6$\times$)}}
& \cellcolor[HTML]{D5E8D4}\textbf{39.34} & \cellcolor[HTML]{D5E8D4}\textbf{2.10} & \cellcolor[HTML]{D5E8D4}\textbf{48.66} \textcolor[HTML]{D35400}{\textbf{(6.2$\times$)}}
& \cellcolor[HTML]{D5E8D4}\textbf{54.91} & \cellcolor[HTML]{D5E8D4}\textbf{1.86} & \cellcolor[HTML]{D5E8D4}\textbf{49.88} \textcolor[HTML]{D35400}{\textbf{(2.9$\times$)}} \\
\hhline{~-----------}

& \multirow{5}{*}{512} & \cellcolor[HTML]{D3D3D3}Baseline
& \cellcolor[HTML]{D3D3D3}14.12 & \cellcolor[HTML]{D3D3D3}1.00 & \cellcolor[HTML]{D3D3D3}6.91 \textcolor[HTML]{D35400}{(1.0$\times$)}
& \cellcolor[HTML]{D3D3D3}41.00 & \cellcolor[HTML]{D3D3D3}1.00 & \cellcolor[HTML]{D3D3D3}5.15 \textcolor[HTML]{D35400}{(1.0$\times$)}
& \cellcolor[HTML]{D3D3D3}53.80 & \cellcolor[HTML]{D3D3D3}1.00 & \cellcolor[HTML]{D3D3D3}10.80 \textcolor[HTML]{D35400}{(1.0$\times$)} \\
& & \cellcolor[HTML]{D3D3D3}Baseline+Parallel
& \cellcolor[HTML]{D3D3D3}14.12 & \cellcolor[HTML]{D3D3D3}2.73 & \cellcolor[HTML]{D3D3D3}18.90 \textcolor[HTML]{D35400}{(2.7$\times$)}
& \cellcolor[HTML]{D3D3D3}39.33 & \cellcolor[HTML]{D3D3D3}1.88 & \cellcolor[HTML]{D3D3D3}9.69 \textcolor[HTML]{D35400}{(1.9$\times$)}
& \cellcolor[HTML]{D3D3D3}53.10 & \cellcolor[HTML]{D3D3D3}1.50 & \cellcolor[HTML]{D3D3D3}30.38 \textcolor[HTML]{D35400}{(2.8$\times$)} \\
& & d$^2$Cache
& 12.29 & 2.89 & 49.76 \textcolor[HTML]{D35400}{(7.2$\times$)}
& 35.48 & 1.62 & 21.65 \textcolor[HTML]{D35400}{(4.2$\times$)}
& 46.00 & 1.93 & 41.11 \textcolor[HTML]{D35400}{(3.8$\times$)} \\
& & EntropyCache
& 13.71 & 2.70 & 49.67 \textcolor[HTML]{D35400}{(7.2$\times$)}
& 37.12 & 1.48 & 20.31 \textcolor[HTML]{D35400}{(3.9$\times$)}
& 47.64 & 1.75 & 39.64 \textcolor[HTML]{D35400}{(3.7$\times$)} \\
& & \cellcolor[HTML]{D5E8D4}\textbf{Polestar (Ours)}
& \cellcolor[HTML]{D5E8D4}\textbf{14.12} & \cellcolor[HTML]{D5E8D4}\textbf{3.06} & \cellcolor[HTML]{D5E8D4}\textbf{51.38} \textcolor[HTML]{D35400}{\textbf{(7.4$\times$)}}
& \cellcolor[HTML]{D5E8D4}\textbf{38.00} & \cellcolor[HTML]{D5E8D4}\textbf{2.17} & \cellcolor[HTML]{D5E8D4}\textbf{46.98} \textcolor[HTML]{D35400}{\textbf{(9.1$\times$)}}
& \cellcolor[HTML]{D5E8D4}\textbf{51.36} & \cellcolor[HTML]{D5E8D4}\textbf{1.93} & \cellcolor[HTML]{D5E8D4}\textbf{46.39} \textcolor[HTML]{D35400}{\textbf{(4.3$\times$)}} \\

\Xhline{2\arrayrulewidth}
\end{tabular}}
\label{tab:appendix_method_comparison_with_baselines}
\end{table*}

~\autoref{tab:appendix_method_comparison_with_baselines} extends the main comparison to d$^2$Cache~\cite{jiang2025d} and EntropyCache~\cite{cheong2026entropycache} across LLaDA-8B-Instruct, LLaDA-1.5, and Dream-7B-Instruct. Evidently, Polestar maintains the strongest accuracy--throughput trade-off across model families and generation lengths. Compared with d$^2$Cache and EntropyCache, Polestar consistently achieves higher TPS and higher TPF while preserving stronger accuracy on reasoning and coding tasks. For instance, on GSM8K with generation length 512, Polestar reaches $80.60$ TPS on LLaDA-8B-Instruct and $61.35$ TPS on LLaDA-1.5, substantially outperforming both dynamic cache baselines. These results support that drift-aware sparse refresh and commitment remain effective beyond the subset of baselines reported in the main table.

\subsection{Multimodal Evaluation with LLaDA-V}
\label{sec:appendix_lladav_results}

\begin{table*}[t]
\centering
\caption{LLaDA-V results on MathVista and MathVerse benchmarks.}
\renewcommand*{\arraystretch}{0.85} 
\setlength{\tabcolsep}{5pt}
\resizebox{0.8\textwidth}{!}{
\begin{tabular}{l | ccc | ccc}
\Xhline{2\arrayrulewidth}
\multirow{2}{*}{Method} & \multicolumn{3}{c|}{MathVista} & \multicolumn{3}{c}{MathVerse} \\
& Acc(\%)$\uparrow$ & TPF$\uparrow$ & TPS$\uparrow$ & Acc(\%)$\uparrow$ & TPF$\uparrow$ & TPS$\uparrow$ \\
\Xhline{2\arrayrulewidth}

\cellcolor[HTML]{D3D3D3}Baseline & 
\cellcolor[HTML]{D3D3D3}60.61 & \cellcolor[HTML]{D3D3D3}1.00 & \cellcolor[HTML]{D3D3D3}2.23 & 
\cellcolor[HTML]{D3D3D3}34.00 & \cellcolor[HTML]{D3D3D3}1.00 & \cellcolor[HTML]{D3D3D3}2.10 \\

\cellcolor[HTML]{D3D3D3}Baseline+Parallel & 
\cellcolor[HTML]{D3D3D3}59.40 & \cellcolor[HTML]{D3D3D3}1.88 & \cellcolor[HTML]{D3D3D3}4.67 & 
\cellcolor[HTML]{D3D3D3}33.20 & \cellcolor[HTML]{D3D3D3}3.71 & \cellcolor[HTML]{D3D3D3}3.34 \\

Fast-dLLM \cite{wu2025fast} & 60.92 & 1.70 & 27.00 & 32.82 & 2.26 & 32.79 \\
Elastic-Cache \cite{nguyen2025attention} & 62.54 & 1.74 & 27.68 & 34.76 & 2.41 & 35.12 \\
d$^2$Cache \cite{jiang2025d} & 62.80 & 1.95 & 23.50 & 35.60 & 2.52 & 26.80 \\
EntropyCache \cite{cheong2026entropycache} & 60.50 & 1.72 & 31.20 & 33.50 & 2.30 & 38.40 \\
Dynamic-dLLM \cite{wudynamic} & 61.80 & 1.78 & 29.50 & 34.20 & 2.38 & 36.20 \\

\cellcolor[HTML]{D5E8D4}\textbf{Polestar (Ours)} & 
\cellcolor[HTML]{D5E8D4}\textbf{63.58} & \cellcolor[HTML]{D5E8D4}\textbf{2.17} & \cellcolor[HTML]{D5E8D4}\textbf{37.91} & 
\cellcolor[HTML]{D5E8D4}\textbf{36.12} & \cellcolor[HTML]{D5E8D4}\textbf{2.85} & \cellcolor[HTML]{D5E8D4}\textbf{42.95} \\

\Xhline{2\arrayrulewidth}
\end{tabular}}
\label{tab:llada_v_horizontal}
\end{table*}

~\autoref{tab:llada_v_horizontal} evaluates Polestar on multimodal LLaDA-V using MathVista and MathVerse. Polestar achieves the best accuracy, TPF, and TPS on both benchmarks. On MathVista, Polestar reaches $63.58\%$ accuracy and $37.91$ TPS, improving over the strongest cache baselines while increasing decoding parallelism to $2.17$ TPF. On MathVerse, Polestar achieves $36.12\%$ accuracy and $42.95$ TPS, again outperforming both static and dynamic cache-update methods. These results suggest that the same drift signal used for text-only dLLMs transfers to multimodal dLLM inference, where newly decoded textual tokens can also induce representation changes in nearby multimodal context.

\subsection{ParallelBench-\textit{Hard} Results}
\label{sec:appendix_parallel_bench}










\begin{table}[t]
\centering
\begin{minipage}{0.55\textwidth} 
\centering
\caption{ParallelBench-\textit{Hard} results. Higher Acc/TPS is better ($\uparrow$), higher TPF indicates more parallelism ($\uparrow$).}
\label{tab:parallelbench_hard}

\renewcommand*{\arraystretch}{1.1}
\setlength{\tabcolsep}{4pt}
\resizebox{0.7\linewidth}{!}{%
\begin{tabular}{l | c | c | c}
\Xhline{2\arrayrulewidth}
Method & Acc(\%)$\uparrow$ & TPF$\uparrow$ & TPS$\uparrow$ \\
\Xhline{2\arrayrulewidth}

\cellcolor[HTML]{D3D3D3}Baseline (LLaDA)
& \cellcolor[HTML]{D3D3D3}86.08
& \cellcolor[HTML]{D3D3D3}1.00
& \cellcolor[HTML]{D3D3D3}6.59 \\

\cellcolor[HTML]{D3D3D3}Baseline+Parallel
& \cellcolor[HTML]{D3D3D3}84.13
& \cellcolor[HTML]{D3D3D3}1.47
& \cellcolor[HTML]{D3D3D3}26.87 \\

Fast-dLLM \cite{wu2025fast}
& 72.13 & 1.33 & 33.18 \\

Elastic-Cache \cite{nguyen2025attention}
& 73.84 & 1.36 & 37.52 \\

d$^2$Cache \cite{jiang2025d}
& 74.62 & 1.39 & 36.14 \\

EntropyCache \cite{cheong2026entropycache}
& 76.15 & 1.42 & 42.85 \\

Dynamic-dLLM \cite{wudynamic}
& 73.21 & 1.35 & 36.19 \\

\cellcolor[HTML]{D5E8D4}\textbf{Polestar (Ours)}
& \cellcolor[HTML]{D5E8D4}\textbf{78.63}
& \cellcolor[HTML]{D5E8D4}\textbf{1.55}
& \cellcolor[HTML]{D5E8D4}\textbf{55.34} \\

\Xhline{2\arrayrulewidth}
\end{tabular}}
\end{minipage}
\end{table}

~\autoref{tab:parallelbench_hard} reports results on the hard subset of ParallelBench~\cite{kang2025parallelbench}, which is designed to stress-test the accuracy--parallelism trade-off. As expected, all cache-enabled and parallel-decoding methods suffer larger accuracy drops than on the easy subset, since premature or overly aggressive commitment is more likely to change the final answer. Nevertheless, Polestar remains the strongest method among the optimized inference baselines, achieving $78.63\%$ accuracy, $1.55$ TPF, and $55.34$ TPS. Compared with EntropyCache, Polestar improves accuracy by $2.48\%$ and increases TPS by $1.29\times$, indicating that drift-aware commitment is more robust under hard-to-parallelize decoding dynamics.

ParallelBench-\textit{Hard} is intentionally adversarial to parallel decoding because later tokens often depend on unresolved earlier reasoning states. Therefore, all accelerated methods show degradation relative to full recomputation. Polestar reduces, but does not eliminate, this degradation.

\begin{table*}[t]
\centering
\caption{LLaDA-8B-Instruct throughput on GH200 across benchmarks and generation lengths.}
\renewcommand*{\arraystretch}{1.1}
\setlength{\tabcolsep}{4pt}
\resizebox{0.6\textwidth}{!}{%
\begin{tabular}{lc|ccc}
\Xhline{2\arrayrulewidth}
Benchmark & Gen Len & Fast-dLLM & Elastic-Cache & \textbf{Polestar (Ours)} \\
\Xhline{2\arrayrulewidth}

\multirow{2}{*}{GSM8K}
& 256 & 77.25 & 85.42 & \cellcolor[HTML]{D5E8D4}\textbf{151.31} \\
& 512 & 58.07 & 64.87 & \cellcolor[HTML]{D5E8D4}\textbf{131.22} \\
\Xhline{1\arrayrulewidth}

\multirow{2}{*}{MATH}
& 256 & 57.97 & 71.40 & \cellcolor[HTML]{D5E8D4}\textbf{105.99} \\
& 512 & 83.89 & 59.79 & \cellcolor[HTML]{D5E8D4}\textbf{106.98} \\
\Xhline{1\arrayrulewidth}

\multirow{2}{*}{HumanEval}
& 256 & 89.34 & 130.18 & \cellcolor[HTML]{D5E8D4}\textbf{134.96} \\
& 512 & 92.90 & 102.08 & \cellcolor[HTML]{D5E8D4}\textbf{142.27} \\

\Xhline{2\arrayrulewidth}
\end{tabular}}
\label{tab:gh200_tps}
\end{table*}
\subsection{Throughput on GH200}
\label{sec:appendix_gh200}
In \autoref{tab:gh200_tps}, we report throughput measurements on an NVIDIA GH200 GPU.
\subsection{Multi-Token Decoding Induces Representation Drift}
\label{sec:appendix_selective_update}

\begin{figure}[t]
    \centering
    \includegraphics[width=0.6\linewidth]{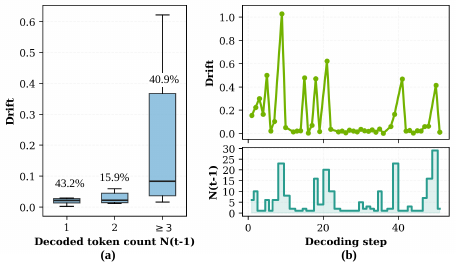}
    \caption{
    Multi-token decoding induces local representation drift.
    (a) Local-window drift grouped by the number of tokens decoded in the previous step; percentages above boxes denote the fraction of transitions in each group.
    (b) A representative trajectory showing that drift spikes align with high decoded-token-count steps.
    }
    \label{fig:burst_drift}
\end{figure}

We validate the update schedule of Polestar-Cache using full-sequence trajectories with confidence-based parallel decoding on LLaDA-8B-Instruct~\cite{nie2025large}. 
For each transition, we associate the previous-step decoded-token count $N^{(t-1)}$ with the ensuing local-window drift measured as $\mathrm{KL}(\mathbf{A}^{(t)}\|\mathbf{A}^{(t-1)})$. 
\autoref{fig:burst_drift}(a) shows that larger previous-step decoded-token counts induce higher subsequent drift, with the strongest increase when $N^{(t-1)}\ge3$. 
\autoref{fig:burst_drift}(b) further illustrates that drift spikes align with high decoded-token-count steps in the trajectory. 
These results support Polestar's decoded-token-count-aware refresh schedule, which reacts to abrupt or accumulated contextual updates rather than recalibrating periodically.
\subsection{Extended Hyperparameter Ablation}
\begin{figure}[t]
    \centering
    \includegraphics[width=0.8\linewidth]{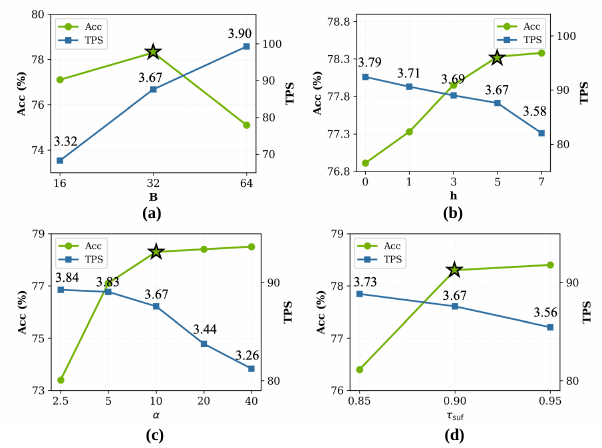}
    \caption{
    Extended hyperparameter ablation.
    (a) Block size $\mathcal{B}$.
    (b) Drift history size $\textbf{h}$.
    (c) Current-block drift gate coefficient $\alpha$.
    (d) Suffix pre-commit confidence threshold $\tau_{\mathrm{suf}}$.
    The default configuration is marked with a star and text annotations indicate TPF.
    Across sweeps, Polestar remains stable near its default setting.
    }
    \label{fig:extended_hparam}
\end{figure}

\textbf{Block size ${B}$. }
We first vary the block size ${B}$ while keeping the generation length fixed to 256.
As shown in~\autoref{fig:extended_hparam}(a), smaller blocks incur more block transitions with higher cache recomputation at block entry, resulting in lower throughput.
Increasing the block size to ${B}=64$ improves TPF and TPS, but substantially degrades accuracy to $75.1\%$, suggesting that larger blocks cause premature commitment risk. The default ${B}=32$ provides the strongest accuracy--throughput trade-off.

\textbf{Drift history size $\textbf{h}$.}
We study the history size $\textbf{h}$ used to compute recent-history-relative drift.
As shown in~\autoref{fig:extended_hparam}(b), using raw drift without history normalization ($\textbf{h}=0$) is aggressive, yielding high TPF and TPS but reducing accuracy to $76.91\%$.
Increasing $\textbf{h}$ stabilizes the drift signal by filtering transient spikes and steadily improves accuracy.
The default $\textbf{h}=5$ provides a strong accuracy--throughput balance, while $\textbf{h}=7$ is more conservative and yields only a marginal accuracy gain at a larger throughput cost.
This supports using relative drift spikes rather than absolute drift magnitude for commitment readiness.

\textbf{Drift gate coefficient $\alpha$.}
We ablate the coefficient $\alpha$ in the confidence-conditioned drift gate.
As shown in~\autoref{fig:extended_hparam}(c), small $\alpha$ values make the gate more aggressive: $\alpha=2.5$ reaches high TPF but sharply reduces accuracy to $73.4\%$.
Larger values are safer but increasingly conservative, lowering TPF and TPS.
The default $\alpha=10$ lies in a stable accuracy--parallelism region, preserving accuracy while retaining most of the speedup from drift-aware commitment.

\textbf{Suffix confidence threshold. }
Finally, we vary the confidence threshold for suffix pre-commitment.
As shown in~\autoref{fig:extended_hparam}(d), lowering the threshold to $0.85$ increases TPF but reduces accuracy to $76.4\%$, indicating unsafe suffix commitment.
Raising the threshold to $0.95$ is slightly more conservative, preserving accuracy but reducing TPF and TPS.
The threshold of $0.9$ which is identical to $\tau_s$ provides the best balance, supporting our design in which drift determines suffix eligibility and confidence determines safe commitment.










\begin{table}[H] 
\centering
\begin{minipage}{0.48\textwidth} 
\centering
\caption{Ablation on quantization format.}
\label{tab:quant_ablation}

\renewcommand*{\arraystretch}{1.1} 
\setlength{\tabcolsep}{6pt}
\resizebox{\linewidth}{!}{
\begin{tabular}{l | c | c}
\Xhline{2\arrayrulewidth}
Quantization Format & Residual Quant. & Accuracy (\%)$\uparrow$ \\
\Xhline{2\arrayrulewidth}

\cellcolor[HTML]{D3D3D3}FP16 (Full Precision)
& \cellcolor[HTML]{D3D3D3}-- 
& \cellcolor[HTML]{D3D3D3}78.90 \\

FP8
& Yes
& 78.90 \\

INT8
& Yes
& 78.54 \\

MXFP4
& Yes
& 78.19 \\

INT2
& Yes
& 0.00 \\

NVFP4
& No
& 54.98 \\

\cellcolor[HTML]{D5E8D4}\textbf{NVFP4 (Polestar)}
& \cellcolor[HTML]{D5E8D4}\textbf{Yes}
& \cellcolor[HTML]{D5E8D4}\textbf{78.33} \\

\Xhline{2\arrayrulewidth}
\end{tabular}}
\end{minipage}
\end{table}
\textbf{Quantization. }We ablate the compression format for Polestar's cached local-window hidden states. As shown in~\autoref{tab:quant_ablation}, FP8 and INT8 remain close to FP16, whereas direct NVFP4 quantization without residuals drops accuracy to $54.98\%$. Residual quantization recovers NVFP4 accuracy to $78.33\%$, closely matching the FP16 baseline while using a much lower-precision representation. We therefore use residual-based NVFP4 by default.

\begin{figure}[t]
    \centering
    \includegraphics[width=\linewidth]{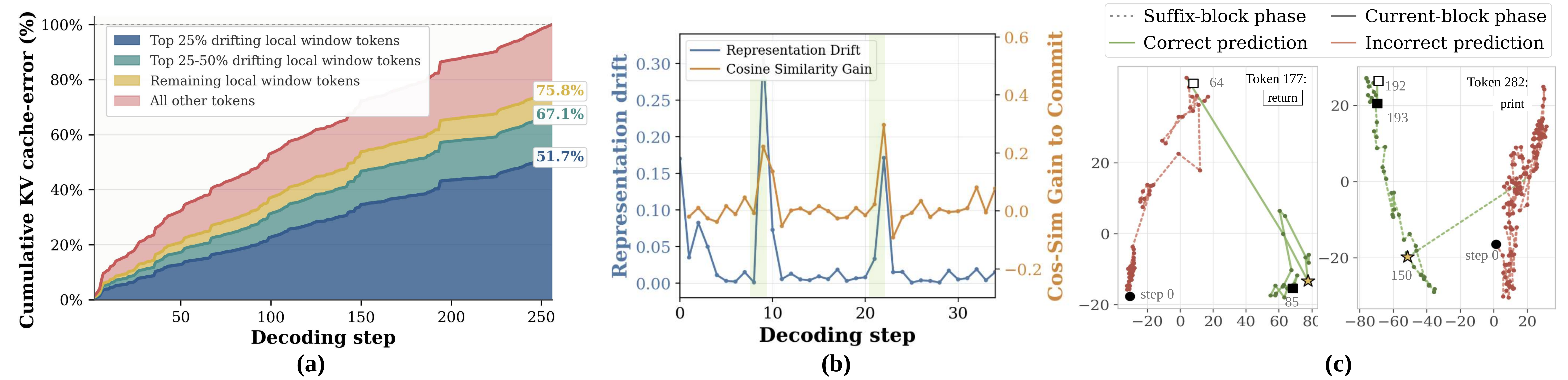}
    \caption{ Additional HumanEval motivation analysis for Polestar:
    (a) KV-cache error concentrates on high-drift states in the local window around the active decoding block;
    (b) sharp drift events coincide with progress toward the token's final predictive state;
    (c) PCA trajectories show that the first drift trigger aligns with the onset of stable prediction trajectories, including suffix-block positions before they become active.
    }
    \label{fig:humaneval_motivation}
\end{figure}

\begin{figure}[t]
    \centering
    \includegraphics[width=0.34\linewidth]{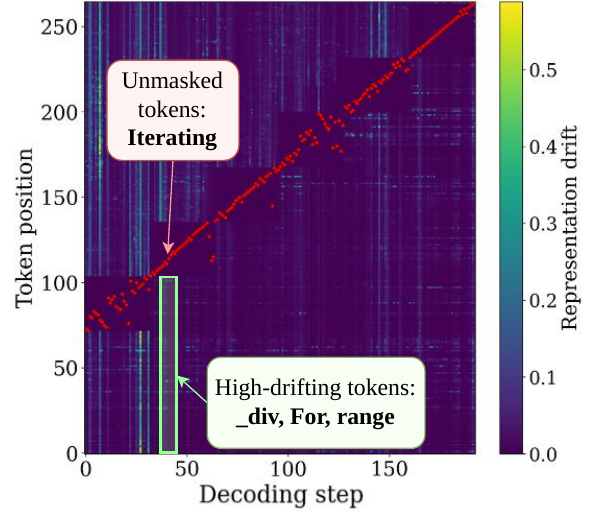}
    \caption{ HumanEval drift heatmap depicting measured drift across token positions and decoding steps, with unmasked tokens shown in red.
    Unmasking a code-relevant token induces drift at related syntactic and semantic positions, illustrating distributed contextual adaptation in code generation.
    }
    \label{fig:humaneval_heatmap}
\end{figure}

\subsection{Extended Motivational Analysis on HumanEval}
\label{app:humaneval_motivation}

We provide additional motivating analyses on HumanEval in \autoref{fig:humaneval_motivation} and \autoref{fig:humaneval_heatmap} to demonstrate the generalization of our observations.
\subsection{Performance Breakdown}
\label{sec:appendix_performance_breakdown}
\begin{figure}[t]
  \centering
  \includegraphics[width=0.9\linewidth, keepaspectratio]{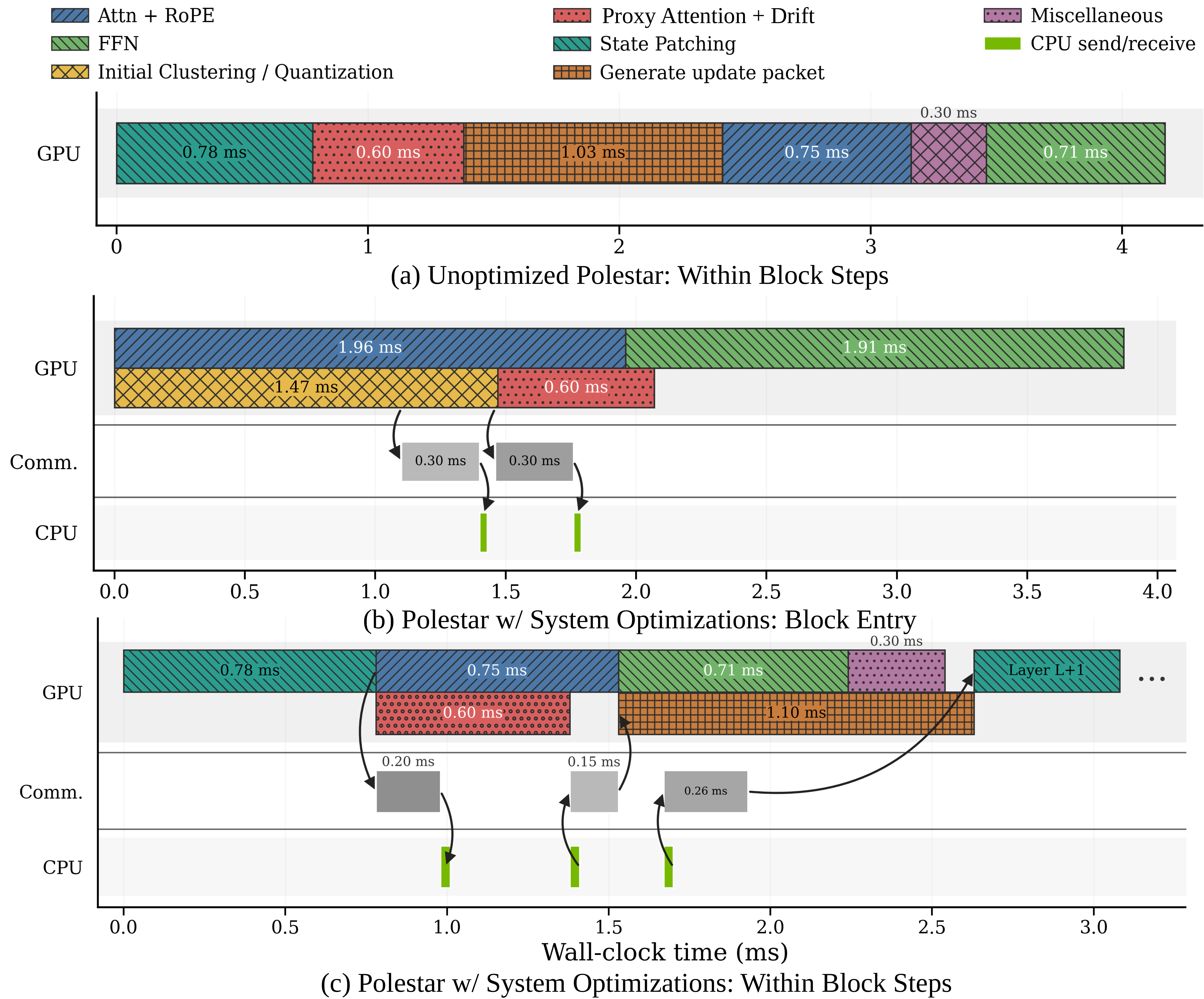}
  \caption{Timing breakdown of Polestar execution. (a) Unoptimized Polestar, (b) Optimized Polestar at block entry and (c) Optimized Polestar at all other steps.}
  \label{fig:performance_breakdown}
\end{figure}

\autoref{fig:performance_breakdown} shows Polestar's worst-case per-layer, per-step timing, where all Polestar-specific operations are active. In the unoptimized execution (\autoref{fig:performance_breakdown}(a)), Polestar-specific operations add $2.71$ ms of critical-path overhead. Since block entry and within-block decoding involve different operations in Polestar, \autoref{fig:performance_breakdown}(b) and (c) show the optimized execution for each case. At block entry, the attention and FFN stages are longer due to the full-sequence forward pass at every alternate block entry, but Polestar-specific operations are fully overlapped with the main computation. For subsequent within-block steps, overlap reduces the added overhead to $\leq0.10$ ms. Polestar's efficient prefetching of next layer's hidden states saves up to $0.26$ms per-step, per-layer. Overall, the proposed system optimizations reduce per-layer latency by $1.63$ ms over the unoptimized case, directly contributing to Polestar's higher throughput.

\begin{figure}[t]
    \centering
    \includegraphics[width=0.5\linewidth]{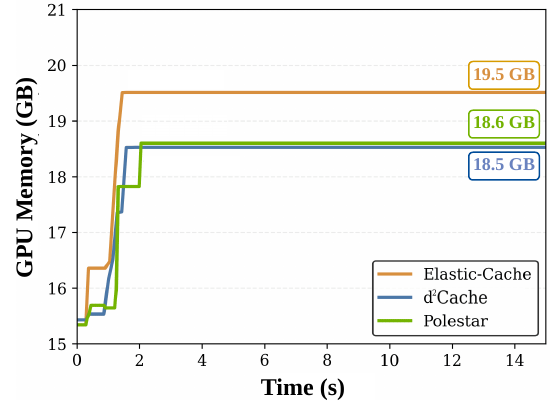}
    \caption{ GPU-memory traces comparison during decoding on LLaDA-8B-Instruct with 5-shot GSM8K, generation length 512, block size 32, batch size 1 on an A100 80GB GPU.
    }
    \label{fig:appendix_mem}
\end{figure}

\subsection{Memory Analysis. }
We compare the inference-time GPU memory footprint of Polestar, Elastic-Cache~\cite{nguyen2025attention}, and d$^2$Cache~\cite{jiang2025d} during generation on LLaDA-8B-Instruct with GSM8K, generation length 512, and block size 32.
For each method, we load the model first and start tracing allocated GPU memory immediately before decoding.
Thus, the reported values include resident model weights, KV-cache buffers, and method-specific runtime buffers, while excluding model-loading time.

As shown in~\cref{fig:appendix_mem}, Elastic-Cache reaches the largest peak memory footprint at $19.5$ GB.
By comparison, d$^2$Cache and Polestar peak at $18.5$ GB and $18.6$ GB, respectively.
Thus, Polestar remains memory-comparable to d$^2$Cache while using $0.9$ GB less peak GPU memory than Elastic-Cache.
This indicates that Polestar's drift-aware cache calibration does not require a large auxiliary GPU-memory footprint: only compact centroids are kept on GPU, while quantized hidden-state residuals for the local window are offloaded to pinned CPU memory.

\section{Declaration of LLM Usage}
\label{sec:appendix_llm_usage}

We used general-purpose LLM assistants only for writing, editing, formatting, and presentation support.
This included grammar correction, concision, LaTeX polishing, table/caption wording, checklist drafting, and identifying possible clarity issues in the manuscript.
LLMs were not used as an important, original, or non-standard component of the proposed method, theoretical claims, experimental design, benchmark evaluation, or reported results.
All technical content, mathematical arguments, implementation details, experiments, numerical results, and final claims were written, checked, and approved by the authors.



\end{document}